\documentclass[journal]{IEEEtran}
\usepackage{amssymb}
\usepackage{amsmath}
\usepackage{multirow, multicol}
\usepackage{makecell}
\usepackage{graphicx}
\usepackage{subfigure}

\usepackage{cite}

\ifCLASSINFOpdf
\else
\fi
\begin{document}
%
\title{Fine--Grained Extraction of Road Networks via \\ Joint Learning of Connectivity and Segmentation}
%
%
%

\author{Yijia~Xu,
		Liqiang~Zhang,~\IEEEmembership{Member,~IEEE,}
		Wuming~Zhang,
		Suhong~Liu, 
		Jingwen Li,
		Xingang~Li, 
		Yuebin~Wang,~\IEEEmembership{Member,~IEEE,}
		and~Yang~Li
\thanks{Y. Xu, L. Zhang, S. Liu, X, Li and Y, Li are with the Faculty of Geography, Beijing Normal University, Beijing 100875, China (e-mail: 201821051185@mail.bnu.edu.cn, zhanglq@bnu.edu.cn, liush@bnu.edu.cn, lixg95@126.com, isliyang@mail.bnu.edu.cn;).}
\thanks{W. Zhang is with School of Geospatial and Engineering Science, Sun Yat-sen University, Zhuhai 519082, China (e-mail:wumingz@bnu.edu.cn).}
\thanks{J. Li is with College of Geomatics and Geoinformation, Guilin University of Technology, Guilin 541006, China (e-mail: lijw@glut.edu.cn). }%
\thanks{Y. Wang is with the School of Land Science and Technology, China University of Geosciences, Beijing 100083, China (e-mail: xxgcdxwyb@163.com).}
}

%
%

\markboth{Journal of \LaTeX\ Class Files,~Vol.~14, No.~8, January~2021}%
{Shell \MakeLowercase{\textit{et al.}}: Bare Demo of IEEEtran.cls for IEEE Journals}
%



\maketitle

\begin{abstract}
Road network extraction from satellite images is widely applicated in intelligent traffic management and autonomous driving fields. The high-resolution remote sensing images contain complex road areas and distracted background, which make it a challenge for road extraction. In this study, we present a stacked multitask network for end-to-end segmenting roads while preserving connectivity correctness. In the network, a global-aware module is introduced to enhance pixel-level road feature representation and eliminate background distraction from overhead images; a road-direction-related connectivity task is added to ensure that the network preserves the graph-level relationships of the road segments. We also develop a stacked multihead structure to jointly learn and effectively utilize the mutual information between connectivity learning and segmentation learning. We evaluate the performance of the proposed network on three public remote sensing datasets. The experimental results demonstrate that the network outperforms the state-of-the-art methods in terms of road segmentation accuracy and connectivity maintenance.
\end{abstract}


%
\IEEEpeerreviewmaketitle

\section{Introduction}
%
%
%
%
\IEEEPARstart{R}{oad} extraction is one of the major research topics in the areas of remote sensing image parsing and autonomous driving. The abundance of remote sensing image data provides an opportunity for accurate road recognition but also poses a challenge due to the cluttered background of the images and the complexity of roads. Fieldwork and manual labeling generate a reliable road network but at the same time require an intensive human workload and time cost. Automatic unsupervised methods have been proposed in \cite{geraud_fast_2004, cheng_urban_2014, doucette_self-organised_2001, cohen_global_1996, kass_snakes_1988}. However, these methods always depend on handcrafted feature descriptors and predefined parameters, which have been proven to have low generalization and low accuracy in complex regions. Recently, deep learning-based supervised methods have achieved high performances. The most common approaches pose road extraction as a segmentation problem \cite{mattyus_deeproadmapper_2017, bastani_roadtracer:_2018, ventura_iterative_2018}, where pixelwise supervision of general semantic segmentation does not consider road connectivity features, resulting in poor topological relationships of road segment predictions. As the connectivity of roads is also important for road extraction, the above segmentation methods usually impose postprocessing to refine disconnected and incomplete road segments \cite{mattyus_deeproadmapper_2017, li_topological_2019, tan_vecroad_2020}. Nevertheless, it is difficult to enforce topological constraints during postprocessing, and multistep approaches do not take advantage of the end-to-end nature of deep learning. \cite{mattyus_deeproadmapper_2017, li_topology-enhanced_2020, newell_stacked_2016, steger_model-based_1995, treash_automatic_2001} iteratively connect road segments by tracing road pixels, but they are not sufficient for maintaining precision alignment.

\begin{figure*}[!t]
	\centering
	\includegraphics[width=\textwidth]{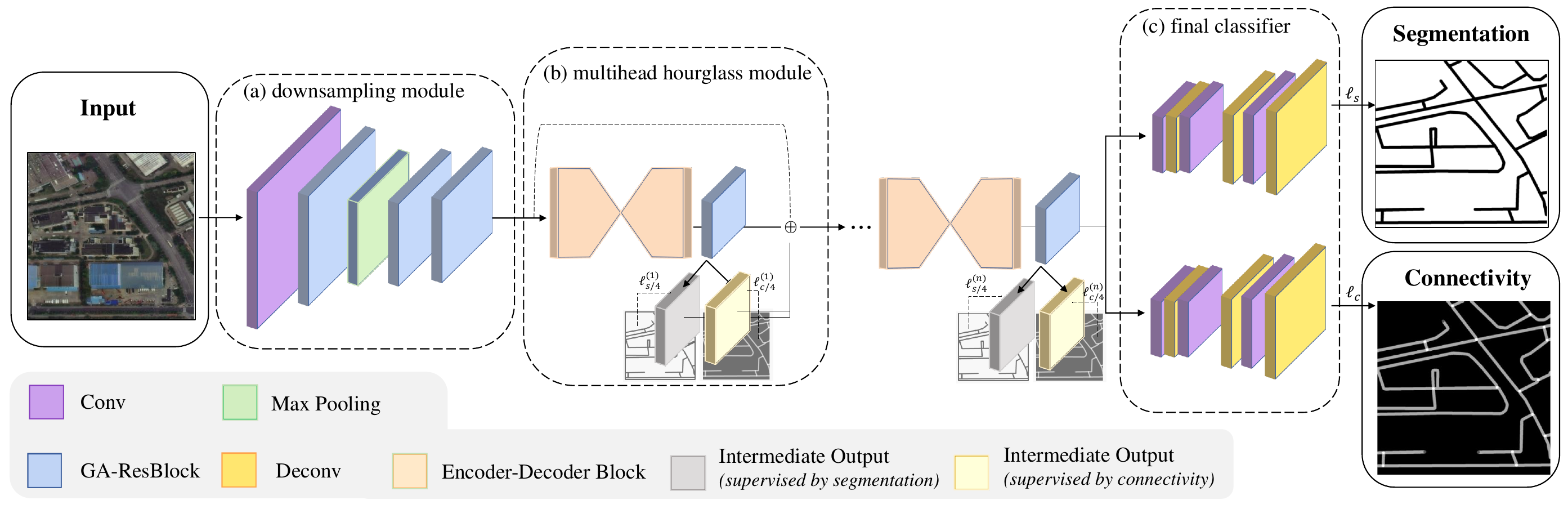}
	\caption{N-stacked multihead architecture, which simultaneously learns road pixel alignment end-to-end and connectivity. Details of the multihead hourglass module are shown in Fig. \ref{fig:MHM}.}
	\label{fig:overall-arch}
\end{figure*}

As a particular semantic segmentation task, remote sensing images pose difficulties in the extraction of roads due to 1) shadows and occlusions of neighboring objects, 2) different appearances due to different materials and illumination, 3) elongation and thinness of road sprawl, 4) spectral similarity between roads and background objects, and 5) intraclass variance in background objects, which also causes false alarms, refers to false positives, where confusing background objects are predicted as roads and inhibits accurate estimation. Considering the complexity of geographical scenes and the imbalance of roads and background in large-scale overhead remote sensing images, we present a deep learning model with global attention that automatically learns road segment connectivity. Global attention is implemented through a global-feature-based attention module, which perceives road-related feature channels and road contextual spatial information to enhance road feature expression and weaken the impact of the noisy geographical environment, thus improving the accuracy of pixel-level road classification. To refine road topological and structural information, we design a task considering connectivity along the road, which conforms to the tracing method for each point when manually annotating road labels. We jointly learn road segmentation along with road connectivity using available labels, which explicitly complement the segmentation features with road connectivity information to generate a topologically connected and pixel-aligned road mask. The full structure of our network, which is a variant of the stacked hourglass for multitask joint learning, is shown in Fig. \ref{fig:overall-arch}. This global-aware multihead stacked network implements topology constraints end-to-end by adding an explicit road connectivity task and takes advantage of road mask feature fusion and road connectivity information through repeated bottom-up, top-down processing. We evaluate the performance of the proposed model on three remote sensing datasets. The experimental results demonstrate that the model outperforms the state-of-the-art methods in terms of segmentation accuracy and connectivity correctness.

Our goal is to enhance road connectivity while maintaining pixel location accuracy. The main contributions are as follows:

\begin{itemize}
\item We propose a stacked multihead end-to-end network to simultaneously predict road segmentation and road connectivity. This architecture automatically fuses the intermediate features of the two tasks to improve the prediction precision of the connected roads and refines multiscale features in repeated U-shaped structures to deal with diverse image resolutions and road scales.
\item A global-aware module is used in our network to replace the normal residual block on account of the complexity of geographical scenes and the thinness and elongation of roads. This lightweight module enables the network to focus on roads and road context, which can enhance road feature expression and suppress background feature interference.
\item In the multihead structure, a novel road-direction-related connectivity task is designed as the auxiliary task of road segmentation. On the one hand, fine road segmentation facilitates road connectivity learning, and connectivity is regressed directly from the masked area with a few misclassified areas. On the other hand, accurate road connectivity refines broken road segmentation, especially at complex intersections.
\end{itemize}

\section{Related Work}

Several methods have been proposed to extract roads from remote sensing images. Some of them tend to utilize handcrafted features to distinguish roads from the background. The imposed connectivity among road points by incorporating geometrical information can generally be classified into edge-based methods \cite{steger_model-based_1995, treash_automatic_2001, ma_road_2007, sengupta_phase-based_nodate, yager_support_2003} and region-based methods \cite{doucette_self-organised_2001-1, park_semi-automatic_2001, chaudhuri_semi-automated_2012-1, shi_integrated_2014}. The former obtains and connects road edge sections and then determines the road regions. The latter aggregates areas with similar grayscale and texture and uses a discriminator to classify road segments. The handcrafted feature descriptors are usually limited to a certain situation and are not robust enough to express complex scenes.

Many recent studies have applied deep learning techniques to road extraction in aerial images. In these techniques, road extraction is generally formulated as a binary semantic segmentation problem in \cite{li_topology-enhanced_2020, zhou_d-linknet:_2018, tao_spatial_2019, liu_d-resunet_2019, sun_stacked_2018}. In our study, we focus on segmentation-based approaches. A neural network technique was introduced to extract roads in aerial images \cite{mnih_learning_2010}. The trained neural network was used to make preliminary predictions and repair gaps as well as disconnected blotches through nearby predictions. In \cite{li_road_2016}, the pixel-level response map was obtained through the CNN, and line integral convolution is developed to connect small gaps. DeepRoadMapper \cite{mattyus_deeproadmapper_2017} initially segmented roads with an encoder-decoder model and repaired missing connections by a linear approximation in the postprocessing to improve the road point connectivity. The following methods focus more on road topological connectivity. RoadTracer \cite{bastani_roadtracer:_2018} used an iterative convolutional neural network (CNN) to predict road graph connectivity, which is not as stable as segmentation-based methods even in areas with visible but dense roads. In \cite{mosinska_beyond_2018}, a higher-order topological loss was utilized to refine the predicted road segments, which was time consuming and inefficient in connecting the road segments in complex scenes. \cite{li_topological_2019} segmented curvilinear structures on the global scale with stacked hourglass \cite{newell_stacked_2016} and estimated the connectivity between points on a local scale by predictions of confidence maps. \cite{li_topological_2019, tan_vecroad_2020} predicted and traced road segments using a CNN or recurrent neural network (RNN). Their performance heavily depends on intersection segmentation results. \cite{batra_improved_2019} added orientation tasks in the road recognition process to help improve road connectivity. \cite{li_topology-enhanced_2020, yu_casenet:_2017, liu_roadnet:_2019} used multitask learning to incorporate road segmentation, where the edge task and centerline task were used as auxiliary tasks. However, the road segmentation task was annotated by a constant width from the road centerline vector. Therefore, the edge label generated from the segmentation label was not accurate, and the centerline task was a repetition of the segmentation task. Since the fusion of multiscale and multilevel features and enlargement of the receptive field may introduce invalid context information, RNN spatial information inference was employed \cite{tao_spatial_2019} to force the network to utilize road-specific contextual information.

UNet \cite{ronneberger_u-net:_2015} and its variants are well used in road segmentation. In these methods, segmentation is generally the first step, and the segmentation results are used to participate in subsequent stages. In \cite{li_feature_2019}, UNet was used to generate road segmentation results before postprocessing. \cite{liu_d-resunet_2019, buslaev_fully_2018, filin_road_2018} combined the ResNet encoder \cite{he_deep_2016} and UNet decoder to detect roads. The assembled multiple UNets \cite{costea_roadmap_2018} and stacked U-Nets \cite{sun_stacked_2018} were utilized to extract roads. The related methods also include D-LinkNet \cite{zhou_d-linknet:_2018} which uses a commonly used segmentation structure LinkNet \cite{chaurasia--linknet_2017} with a dilation convolution center part. In these works, road connectivity is achieved using the connected-tube marked point process \cite{li_feature_2019}, linearly approximated with a fixed length \cite{sun_road_2019}, smoothing-based optimization \cite{costea_roadmap_2018}, and spoke wheel \cite{yujun_end--end_2018}. In contrast, our network directly adds connectivity constraints to the road segmentation process through connectivity task learning, and end-to-end learns the connected road segmentation.

\section{Methodology}

Roads generally span the whole remote sensing image. Their elongation, thinness, and connectivity make traditional semantic segmentation methods generate fragmented road segments. To solve this problem, we propose an end-to-end deep learning framework (Fig. \ref{fig:overall-arch}) to extract roads with connectivity information from remote sensing images. In this framework, a global-aware module is introduced to learn road contextual features of roads, and a road direction-related connectivity task is used to capture the relationship between connected segments by explicitly learning the connectivity among road pixels. Here, we formulate the regression for the road connectivity task as a classification task.

\subsection{Global-Aware Module}\label{sec:GA}

\begin{figure}[!t]
	\centering
	\includegraphics[width=\columnwidth]{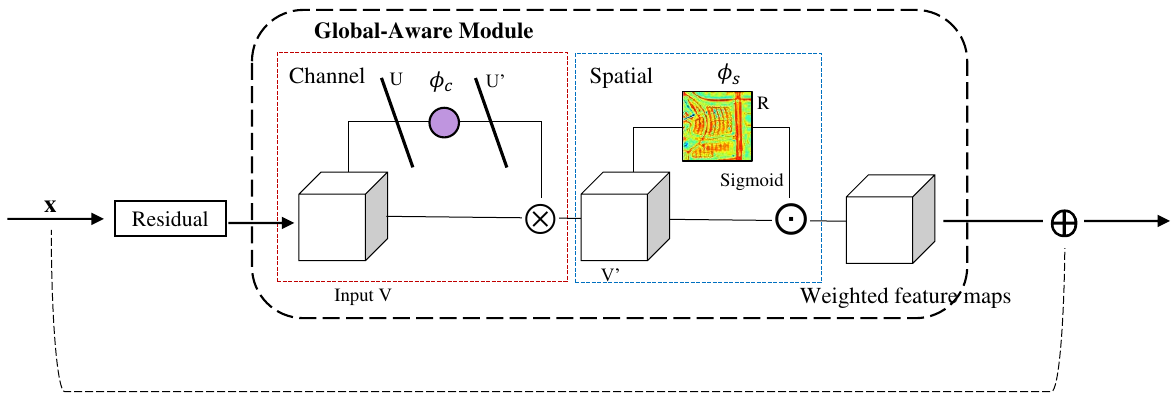}
	\caption{\textbf{GA-ResBlock}. The components in the box make up the global-aware module (\textbf{GA}). GA-ResBlock is obtained by embedding GA into the residual block.}
	\label{fig:ga}
\end{figure}

Global contextual information helps to enhance long-span road feature expression and reduce background false alarms. Considering the road elongation and the cluttered background in the images, we introduce a feature enhancement module with the channel and spatial global information, called the global-aware module (Fig. \ref{fig:ga}), to enable the network to focus on road-related information in the intermediate feature layers. This module extends from the squeeze-and-excitation module (SE module) in SENet \cite{hu_squeeze-and-excitation_2019} by adding ignored global spatial information. The channel attention mechanism in the SE module attains channel importance through global average pooling, which compresses spatial information by equally weighting all pixels in a feature layer. This is a loss of spatial structure that is important for segmentation. Thus, in our framework, we cascade a spatial attention module after a channel attention module, allocate attention on the semantic level and spatial level, and add them in ResBlocks to improve the intermediate feature maps. The improved ResBlock with global attention is called GA-ResBlock (Fig. \ref{fig:ga}). Now, we introduce channel-aware, spatial-aware, and integrated GA-ResBlock.

The channel-aware module is in the red box in Fig. \ref{fig:ga}. It helps attain global receptive fields through adaptive global average pooling, obtaining a vector representing global geographic features. The input feature map is weighted by this vector to enlarge the gap between road-related and background-related feature layers. Details of implementation are as follows: for any input feature map $V = [v_1, v_2, \dots, v_C]$, where $v_i \in \mathbb{R}^{C \times W \times H}$, and $C$ represents the number of channels. To obtain the global information, the channel attention module implements adaptive global average pooling for $V$ to obtain a 1-D scene embedding vector $U\in \mathbb{R}^C$. To further learn the global contextual feature to determine the channel importance, the scene embedding vector $U'$ is computed by applying $\phi_c (\cdot)$ on $U$, as shown in Eq. \eqref{eq:ca},
\begin{equation}\label{eq:ca}
	U' = \phi_c(U)
\end{equation}
where $\phi_c$ denotes a projection function for geospatial scene representation, and it is implemented by two $1 \times 1$ convolution layers to learn the correlation of each channel and then normalized by a sigmoid gate function with an output vector, that is, the weight of each channel $U' \in \mathbb{R}^C$ used for channel weighting. To simplify computation and reduce parameters, the hidden layer of $\phi_c$ reduces dimensionality with a reduction ratio $r$. Then, the input feature map $V$ is multiplied by the scene embedding vector $U'$ to obtain the channel attention-enhanced feature map $V'$. Compared with $V'$, $V$ focuses more on the feature layers of road contexts, such as road orientation, road segment connection, and the relationship between road and the surroundings.

The spatial-aware module (in the blue box in Fig. \ref{fig:ga}) is cascaded after the channel-aware module and is a supplement to the channel information. It compresses the global channel information to obtain the spatial relationship map, describing the spatial relationship between roads and their neighborhood. This operation adds spatial information, which is ignored by the pooling operation, back to the model. Details of implementation are as follows: based on the channel attention-enhanced feature map $V'$, the spatial attention map $R \in \mathbb{R}^{1\times W\times H}$ is obtained by the parameter-free operation as shown in Eq. \eqref{eq:sa}.

\begin{equation}\label{eq:sa}
	R=\phi_s(V)=\mbox{mean}(V', \mbox{dim}=1)
\end{equation}

Corresponding to the channel-aware module, information along the channel dimension is aggregated by summing the values of each channel at a given spatial position, and a 2-D feature map $R$ is obtained, where $R_{ij}$ denotes the importance of pixel $(i, j)$. Here, if a convolution layer was used to reencode the spatial relationship map, the size of the input image would need to be fixed, which reduces the network flexibility. $\phi_s$ is a linear operation; if we multiply $V'$ with $R$ directly, multiplication of features before and after the linear function results in feature degradation. To avoid this, an extra sigmoid function is added to introduce nonlinear features and obtain the final spatial attention map $R$, which is used to reweight $V'$ with the output spatial enhanced feature map $V''$. This feature map focuses more on key spatial areas related to roads and replaces the input feature to participate in subsequent operations.

We integrate the global-aware module with ResBlock to obtain GA-ResBlock, which is used in our network for feature extraction and enhancement. To be more specific, the global-aware module is embedded in the residual learning branch to reweight the residual feature maps. The latent global information is used to increase the disparity between roads and background features, thus improving the discrimination of road representation and reducing background distraction.

\subsection{Road Connectivity Task}\label{sec:Conn}

As pixel-level road annotation is time consuming with data redundancy, road labels are generally vector formats consisting of vertices and their connection relationships. Segmentation-based road extraction requires binarized pixel labels, which means that most of the topology information is lost and that there are omission and registration noise in road ground truths \cite{mnih_machine_2013
}. Road connectivity plays a key role in road extraction. To explicitly learn road connectivity, inspired by ConnNet \cite{kampffmeyer_connnet_2019}, we develop a road direction-related connectivity task that forces the network to count the number of branches emitted by each pixel, thus imposing connectivity constraints on the feature encoding process.

\begin{figure}[!t]
	\centering
	\includegraphics[width=\columnwidth]{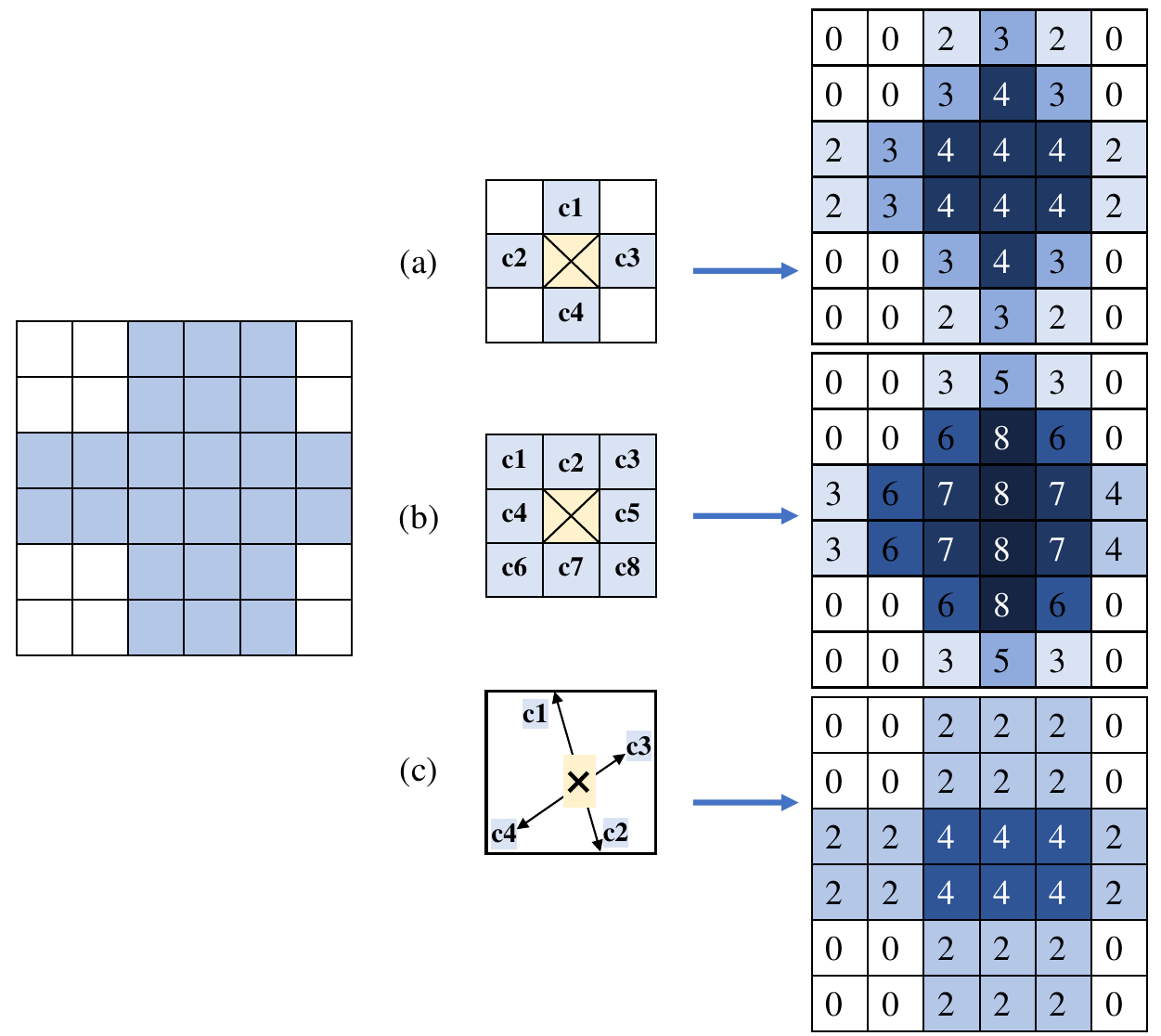}
	\caption{Illustration of different connectivity representations. The left panel is a simplified crossroad grid, where the blue area denotes road pixels and the white area represents the background. The right shows how 4-connectivity (a), 8-connectivity (b), and our road-direction-related connectivity (c) are modeled and how the ground truth is generated. (a) and (b) predict the neighboring four or eight pixels, respectively, for a given pixel. (c) predicts the adjacent pixel along the road. Specifically, four branches are connected at the crossroad, three are connected at the fork, two are connected along the road, and only one is connected at the endpoint. The connectivity label, specifically, is generated by counting the number of connected pixels in the given directions, where a darker color indicates a larger value.}
	\label{fig:conn}
\end{figure}

ConnNet mainly uses 4-connectivity and 8-connectivity, as illustrated in Fig. \ref{fig:conn} (a) and (b). For two pixels $P=(x, y)$ and $Q=(u, v)$, 4-connectivity uses the Manhattan distance as a metric, defined as $d_4(P,Q)=|(x-u)|+|(y-v)|$. For the 8-connectivity, the Chebyshev distance is used, $d_8(P,Q)=\max(|((x-u),|(y-v)|)|)$. They both take pixels as the minimum unit. In road scenes, the connection of two points is expressed as linear distance or Euclidean distance, $d=\sqrt{((x-u)^2+(y-v)^2 )}$. Therefore, it can effectively reflect the topological relationship of road segments if the direction of road linestrings is considered in terms of connectivity (\ref{fig:conn}c).

For a given connectivity pattern, if a pixel and its corresponding neighboring pixels are roads, they are regarded as connected. In this way, the connectivity label is obtained by counting the number of connected pixels for each pixel. For a connectivity map $P$, as shown in Fig. \ref{fig:conn}, $P_{ij}$ denotes the number of neighboring road pixels in a given pattern. As the 4-connectivity and 8-connectivity only consider pixel connectivity, a connected component appears deep inside and shallow at the edges. This only represents the salience of road segments rather than road topological connectivity. In the proposed direction-related connectivity, visualization is darker at the intersection with more branches, which intuitively represents the topological relationship of the road. Road pixels normally have up to five connected directions, that is, $C\leq5$ represents all road connectivity situations.

\begin{figure}[!t]
	\centering
	\includegraphics[width=\columnwidth]{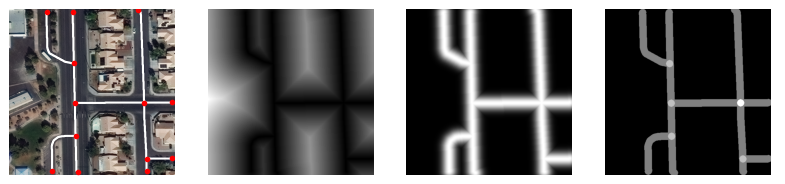}
	\begin{minipage}{\columnwidth}
		\footnotesize
		\hfil
		(a) \hspace{1.7cm}
		(b) \hspace{1.7cm}
		(c) \hspace{1.7cm}
		(d) \hfil
	\end{minipage}
	\caption{Process of generating the road connectivity labels. (a) Road linestrings, (b) a distance map representing the distance from other pixels to the nearest positive pixels. (c) A Gaussian heatmap and (d) the connectivity ground truth. Note that data augmentation, such as cropping, will be applied in the training process, and the node on the boundary is discarded and not treated as an endpoint.}
	\label{fig:conn_pro}
\end{figure}

In the following, we describe the process for generating connectivity labels from the vector ground truths. As shown in Fig. \ref{fig:conn_pro}, road labels are made up of road line strings $\{l_1,l_2,\dots,l_m \}$ and road vertices $\{p_1,p_2,\dots,p_m \}$. For each node, we obtain the number of line segments connected to it, denoting $\{\mbox{deg}_1, \mbox{deg}_2,\dots, \mbox{deg}_m\}$. The road connectivity label is calculated as follows. First, we assign the road pixels as 2 according to the definition in Fig. \ref{fig:conn}. The road line strings are vectorized to obtain the road centerlines. Then, the distance map (Fig. \ref{fig:conn_pro} (b)) is calculated by Eq. \eqref{eq:d}, which is used to calculate the linear distance from a nonroad pixel p0 to its nearest road pixel $p_1$. According to the distance map, Eq. \eqref{eq:G} is used to generate a Gaussian distribution $G\in[0,1]$ (Fig. \ref{fig:conn_pro} (c)). For connectivity label $C$, the pixel where $G_{ij}$ is less than the threshold is assigned $c_r$; for all other pixels, nonroad value 0 is assigned (Eq. \eqref{eq:G_ass}). Then, we reassign the connectivity value of intersections $p_i$ in the same way. By this time, $c_r$ is the number of connected line strings $\mbox{deg}_i$ of this node.
\begin{equation}\label{eq:d}
	d(p_0, p_i) = \|p_0(x, y)-p_i(x, y)\|_2^2
\end{equation}
\begin{equation}\label{eq:G}
	G_{ij}=\exp \left(-\frac{d_{ij}^2}{2\theta^2}\right) \\
\end{equation}
\begin{equation}\label{eq:G_ass}
	G_{ij}=\begin{cases}
		c_r \quad &\mbox{if} \ G_{ij}<\lambda \\
		c_b \quad &\mbox{otherwise.}
	\end{cases}
\end{equation}
where $\|p_i-p_j\|^2_2$ is the Euclidean distance between two points,$(x, y)$ is the pixel coordinates of a point, $G$ is the Gaussian distribution map of the road centerline, and $C$ is the connectivity label.

\subsection{Multihead Hourglass Module}\label{sec:MHM}

\begin{figure}[!t]
	\centering
	\includegraphics[width=\columnwidth]{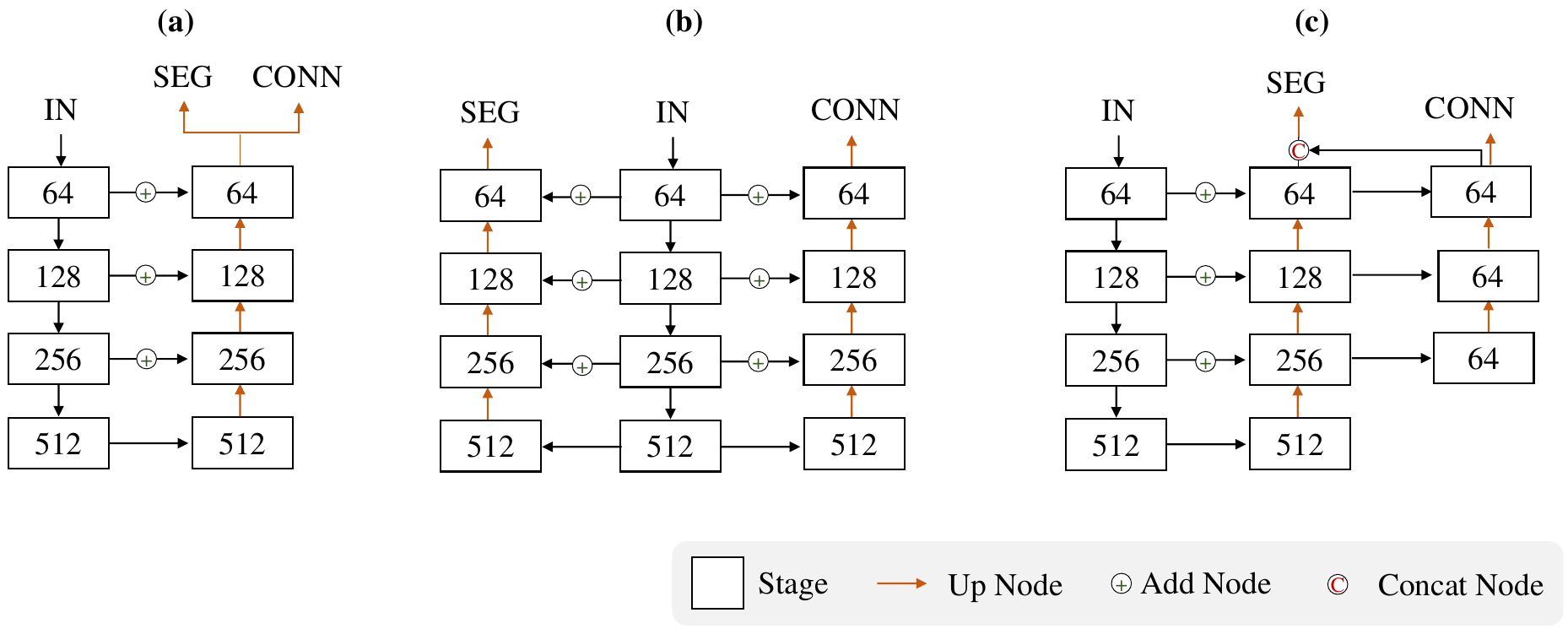}
	\caption{Multitask learning model diagrams. The numbers in the boxes represent the number of channels of the output feature map. (a) is a multihead structure that shares the underlying encoding-decoding network. As an auxiliary task of segmentation, connectivity is then predicted directly from image features at the segmentation coverage. (b) is a multibranch structure. This architecture shares the encoding part, where the network learns robust general feature expressions of connected road segments. (c) is a side-output structure. The side blocks of the decoder network are used for information transformation (segmentation to connectivity) and feature compression. An FPN-like \cite{lin_feature_2017} module is used for upsampling fusion plus a final classification layer for connectivity prediction. We also transmit this captured connectivity information to the backbone road segmentation branch for information enhancement.}
	\label{fig:MHS}
\end{figure}

\begin{figure}[!t]
	\centering
	\includegraphics[width=\columnwidth]{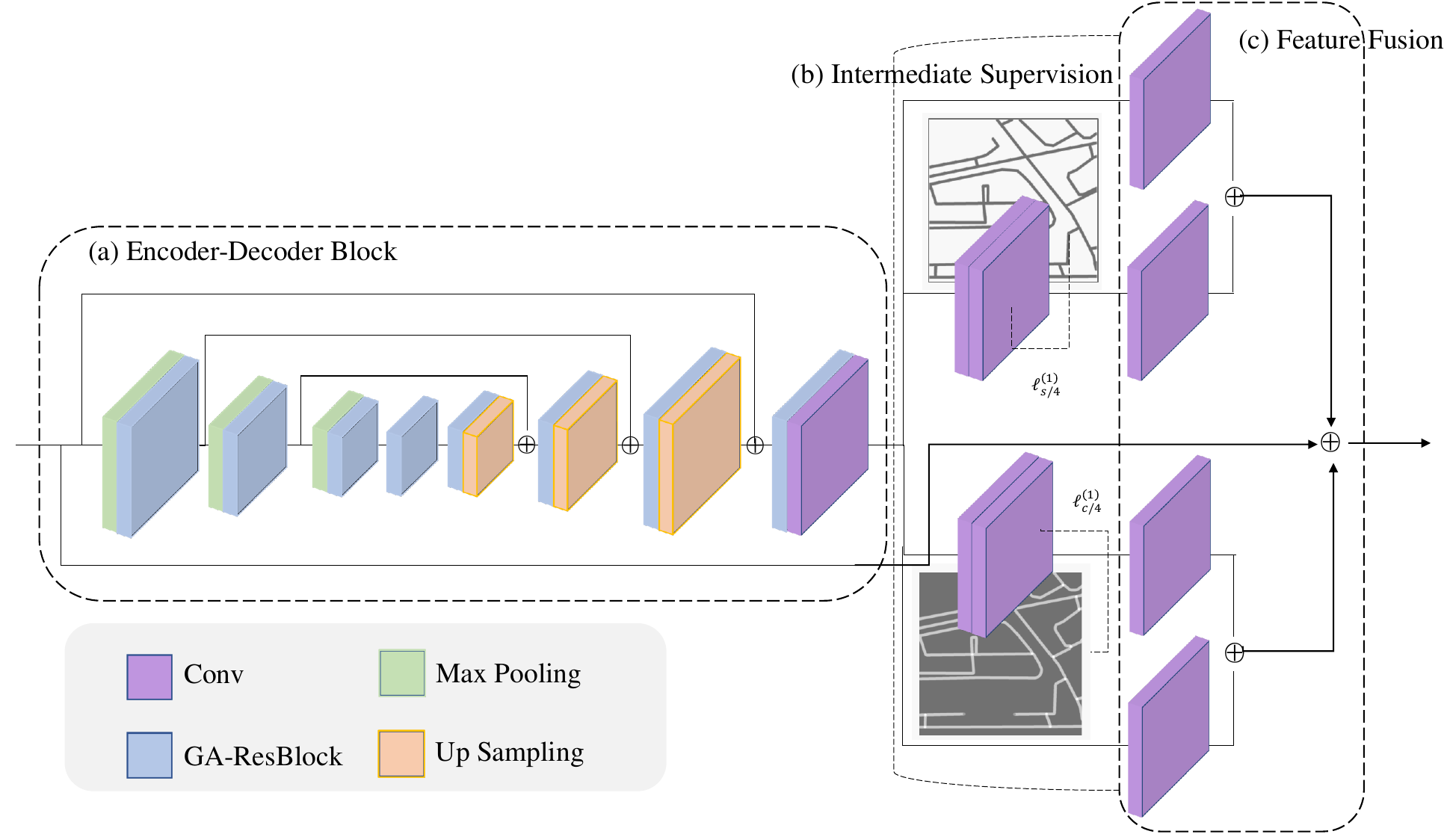}
	\caption{Multihead hourglass module. The dual-head networks have separate prediction and fusion modules. The intermediate output is obtained by a $3\times 3$ feature conversion layer and a $1\times 1$ pixel classification layer, with/4 of the original image size.}
	\label{fig:MHM}
\end{figure}

To integrate the connectivity task and the segmentation task, we use the multihead hourglass module in Fig. \ref{fig:MHM}, which is a variant of the hourglass unit in the stacked hourglass network \cite{newell_stacked_2016}. When determining the network structure, we consider three combination methods: multibranch structure, multihead structure, and side-output structure (taking LinkNet34 \cite{chaurasia--linknet_2017} as an example, the three architectures are shown in Fig. \ref{fig:MHS}). The experiments validate that the multihead network can achieve higher performance with less computational redundancy and is time consuming. We provide more numerical results in Section~\ref{sec:conn_com}.

Fig. \ref{fig:MHM} shows the multihead hourglass module used for stacking (Fig. \ref{fig:overall-arch} (b)). In this module, we introduce a GA-ResBlock in Section~\ref{sec:GA} to obtain global information and enhance the difference between road and nonroad features. The road connectivity task in Section~\ref{sec:Conn} is used as an auxiliary task for segmentation to simultaneously learn robust and generalizable connected road features. In the ith multihead hourglass module, the input feature map $I_i$ passes through a nested U-shaped architecture with three downsamplings and three upsamplings. In the downsampling process, the pooling layers are used to expand the receptive field, reduce feature redundancy, and obtain multiscale spatial and channel information. In the upsampling process, the multiscale features are fused by a skip connection. Different from the original stacked hourglass, we discard additional convolution layers at the skip layer connections for preserving spatial details and instead directly transmit the encoder output identically. The output of this U-shaped network is denoted as $y_i'=U_i (I_i)$, where $i\in[1,2,\cdots,n]$, and $n$ is the number of stacked modules. Then, a GA-ResBlock is used for feature fusion to obtain $y_i\in\mathbb{R}^{d\times H\times W}$. The intermediate supervision block (Fig. \ref{fig:MHM} (b)) consists of a $3\times 3$ feature transformation layer and a $1\times 1$ prediction layer, obtaining 1/4-size coarse segmentation prediction $\overline{s_i} \in \mathbb{R}^{d\times H\times W}$ and connectivity prediction $\overline{c_i} \in \mathbb{R}^{t_2\times H\times W}$ supervised by 1/4-size labels, where t1 and t2 are the numbers of classes of the two tasks. Details about stacking and feature fusion are described in Section~\ref{sec:smn}.

\subsection{Stacked Multihead Network}

The stacked multihead network, as shown in Fig. \ref{fig:overall-arch}, adjusts the stacked hourglass network \cite{newell_stacked_2016} into multitask form. The stacked hourglass has a strong ability to integrate local and global information in the repeated bottom-up and top-down process, while our purpose of using a stacked architecture is to promote information flow between the relevant tasks through the feature fusion process in Fig. \ref{fig:MHM} (c). Our network is composed of three blocks: a downsampling module, an iterative multihead hourglass module, and a pixel-level classifier. The network makes use of the complementary information in the joint learning of the segmentation and connectivity tasks to enhance topological description and segmentation outputs.

The stacked multihead network takes the remote sensing image $X$ as input and then feeds it into the downsampling module (Fig. \ref{fig:overall-arch} (a)). The 1/4-size feature map is obtained by $d=D(X)$, where $D(\cdot)$ represents the downsampling function. $d$ is the input feature ($I_1$) of the first multihead hourglass module. The specific process in the stacked multihead hourglass module is described in Section~\ref{sec:MHM}, where repeated feature learning (obtaining $y_i$) and intermediate supervision (obtaining $\overline{s_i}, \overline{c_i}$ take place. This process is repeated $n$ times. As $i < n - 1$, we fuse $\overline{s_i}, \overline{c_i}, y_i$ by four separate feature transformation layers to project them into the same feature space $\mathbb{R}^{d\times H\times W}$. The transformed features $s'_{i-1}, c'_{i-1}, y_{i-1}^1, y_{i-1}^2$ and input feature $I_i$ are added as an input to the next multihead module. The intermediate output procedure is expressed by Eq. \eqref{eq:sici}.
\begin{equation}\label{eq:sici}
	\overline{s_i}, \overline{c_i} = \begin{cases}
		M_i\left( I_i + s'_{i-1} + c'_{i-1} + y'_{i-1}\right) & if\ i>1 \\
		M_i(I_i) & if\ i=1 
	\end{cases}
\end{equation}
where $M_i(\cdot)$ denotes the multihead module function, $I_i$ is the backbone input feature for each $M_i(\cdot)$, $\overline{s_i} \mbox{ and } \overline{c_i}$ are segmentation and connectivity intermediate outputs, and $s'_{i-1}, c'_{i-1}, y'_{i-1}$ are the projection results of the last multihead decoder features and outputs, where $y'_{i-1}$ includes $y_{i-1}^1 \mbox{ and } y_{i-1}^2$ of two branches separately.

\subsection{Loss Function}

In the segmentation task, we use IoU as the metric. The soft IoU in Eq. \eqref{eq:ls} is taken as the loss function, ensuring that the training process is stepping in the IoU increasing direction. Since the soft IoU is the ratio within each class, the difference in the sample size of each class is irrespective, and class imbalance is avoided.
\begin{equation}\label{eq:ls}
	\ell_s = -\frac{1}{C}\sum_c\sum_{i=1}^h\sum_{j=1}^w\frac{\hat{Y}_{ij}\cdot Y_{ij}}{\hat{Y}_{ij}+Y_{ij}-\hat{Y}_{ij}\cdot Y_{ij}}
\end{equation}
where $\hat{Y}_{ij}$ represents the ground truth for segmentation, and $Y_{ij}$ is the prediction.

For the connectivity task, we use a balanced cross-entropy loss (Eq. \eqref{eq:lc}). Since road intersections account for a tiny part of road pixels, the class size is extremely unbalanced. Therefore, each category is multiplied by a category weight to avoid ignoring a class with a small sample size.
\begin{equation}\label{eq:lc}
	\ell_c=-\frac{1}{\sum w_c}\sum_{c=1}^C \hat{Y}_c \log (Y_c)\cdot w_c
\end{equation}
where $C$ is the number of connectivity classes. Class weights adopt the inverse boundary weight \cite{paszke_enet_2016}, and $w_c=  \frac{1}{\ln(1.02+p_c)}$, which can allow smaller categories to gain more weights. We only expect to adjust the loss distribution without changing the sum of the loss, so we normalize the weighted cross-entropy loss by dividing the number of weights.

Each of the multihead modules produces intermediate supervision, and full-resolution supervision is performed. The total loss is the sum of all losses, as shown in Eq. \eqref{eq:lt}.
\begin{equation}\label{eq:lt}
	\ell = \sum_{i=1}^{n+1}\left(\ell_s^i + \ell_c^i\right)
\end{equation}
where $n$ is the number of stacks.

\section{Implement Details}

To prevent overfitting, we crop the image data into $256\times256$ image patches during training. Additionally, random color jittering, random horizontal flipping, random vertical flipping, and random rotation for $90\times k\ (k=1, 2, 3)$ with augmentation probability $0.5$ are employed for data augmentation. We choose Adam as our optimizer with a weight decay of 5e-4 to prevent overfitting. We train 120 epochs with a cosine annealing learning rate decay, and the initial learning rate is set to 1e-3 and multiplied by $0.5*(1+\cos(\frac{\mbox{step}}{\mbox{max\_step}}\pi))$ to reduce to zero at the end of the training process. The batch size is set to 32 for the training and 1 for the validation and testing. The connectivity category is set to 5. The residual block used in our network is BasicBlock in ResNet-34 with our GA module, and the number of features is set to 128. The number of stacks of our stacked multihead network is set to 2 after the experiment.

For the connectivity measurement, we vectorize the segmentation map into lines. Following \cite{mattyus_deeproadmapper_2017}, we simplify the graph with the Ramer-Douglas-Peucker algorithm, obtaining the approximate road skeleton. To keep the road shape correctly, the tolerance threshold is set to 2 pixels for the SpaceNet dataset. Considering that uneven segmentation results may result in many burrs in the skeletonization process, we remove the short hanging curves of less than 30 pixels from the segmentation results.

\section{Experiment and Discussion}

\subsection{Datasets}
We perform our experiments on three publicly available road datasets, the SpaceNet Road Dataset \cite{van_etten_spacenet:_2019
}, Massachusetts Road Dataset \cite{mnih_machine_2013
} and RoadTracer Dataset \cite{bastani_roadtracer:_2018}, and follow the data preparation process of \cite{batra_improved_2019}. We train, evaluate, and report on images with full resolution. 

\textbf{SpaceNet}: This dataset contains 2,772 remote sensing images from 4 cities with a ground resolution of 0.3 m/pixel and pixel resolution of $1,300\times 1,300$. Annotations are in vector format, representing the centerline of the roads. In the experiment, we split the dataset according to 6:2:2 and obtained 1,523 training images, 514 validation images, and 512 test images. To augment the training dataset, we crop images into $512\times 512$ image patches with stride 215. Intact patches with good quality are retained, providing $\sim$19K training images. For validation and testing, we used the same crop size without overlapping.

\textbf{RoadTracer}: This dataset includes 300 images from 40 cities with a ground resolution of 0.6 m/pixel and pixel resolution of $4,096\times 4,096$. The centerline ground truth is obtained from the Open Street Map (OSM) \cite{noauthor_openstreetmap_nodate}. According to the same ratio, we obtain 181 training images, 58 validation images, and 61 test images, which are also cropped into $512\times 512$ patches, where stride 256 is used to augment training data, obtaining $\sim$33 K.

\textbf{Massachusetts}: This dataset provides 1,171 aerial images with at least $1,500\times 1,500$ pixel resolution and 1.2 m/pixel ground resolution. The corresponding road maps are generated by rasterizing OSM data into lines with a thickness of 7 pixels. We divided the dataset into 687, 240, and 244 parts for training, validation, and testing, respectively. Then, we crop them into $512\times 512$ image tiles; in the same way, the training data overlap with stride 256, obtaining $\sim$12 K.

\subsection{Metrics}

To assess the quantitative and qualitative performance in both road surface segmentation and road centerline connectivity, pixel-based metrics and graph-based metrics are introduced.

Segmentation Metrics: The road segmentation task applies to the commonly used pixel-based segmentation metric: the intersection of unions (IoU). However, the ground truth of road segmentation is generated from centerline line strings with constant width, which means that pixel labels are only accurate up to a few pixels. Therefore, we adopt the relaxed mechanism recommended by Mnih \cite{mnih_machine_2013}, which treats predicted road pixels within $\rho$ pixels of the ground-truth value as true positives. Recall is the fraction of true positives among predicted roads, and accuracy is the fraction of true positives among true roads. We use the standard IoU and relaxed IoU ($IoU^r$) as pixel-level metrics.

Connectivity Metrics: To evaluate the connectivity preservation of road predictions, we use the graph-based metric APLS \cite{van_etten_spacenet:_2019} (average path length similarity). APLS is a Dijkstra shortest path algorithm-based graph theory metric that is used to measure the correctness and connectivity of the predicted road graph. This is calculated by summing the difference in the corresponding optimal path lengths between nodes in the ground-truth graph and proposal graph using Eq. \eqref{eq:mtapls}.
\begin{equation}\label{eq:mtapls}
	M_{apls}=1-\frac{1}{N}\sum\min \left\{1, \frac{|L(a, b)-L(a', b')|}{L(a, b)}\right\}
\end{equation}
where $N$ denotes the number of unique paths in an image and $L(a,b)$ is the length of path $(a,b)$. We snap nodes in the ground truth onto a proposal to measure the similarity between two graphs, which is represented as $M_{g\to p}$. Additionally, to penalize false alarm road proposal, symmetric inverse operation, proposal to ground-truth node snapping $M_{p\to g}$, is also required. The final APLS is the mean value of $M_{g\to p}$ and $M_{p\to g}$, using Eq. \eqref{eq:apls}.
\begin{equation}\label{eq:apls}
	APLS=\frac{1}{N}\sum \frac{1}{\frac{1}{M_{g\to p}}+\frac{1}{M_{p\to g}}}
\end{equation}
Here $N$ is the number of images.

\subsection{Performance Validation}
\subsubsection{\textbf{Global-Aware Module}}
\begin{figure}[!t]
\centering
\subfigure[]{
	\begin{minipage}{0.2\columnwidth}
		\includegraphics[width=1\columnwidth]{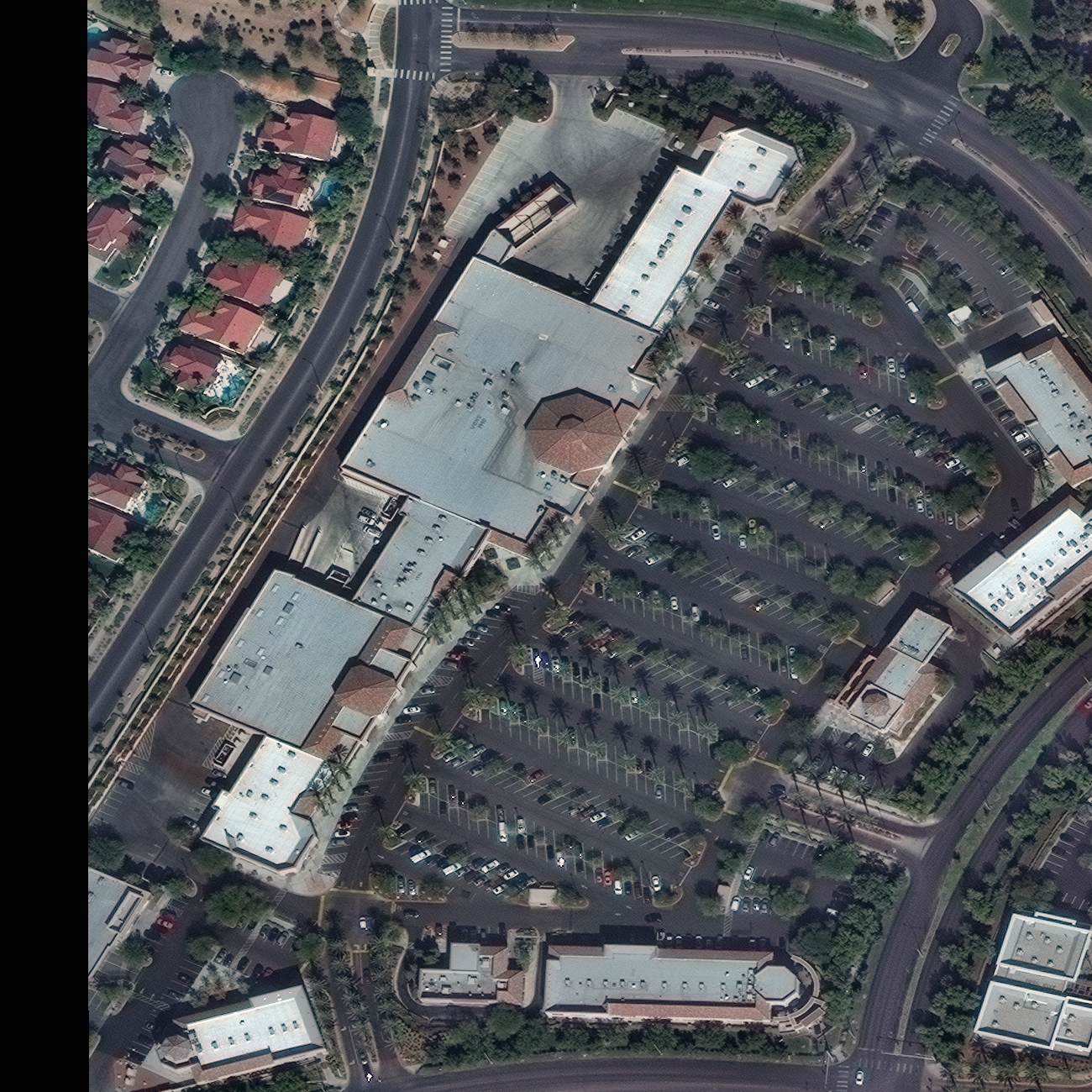} \\
		\vfill
		\includegraphics[width=1\columnwidth]{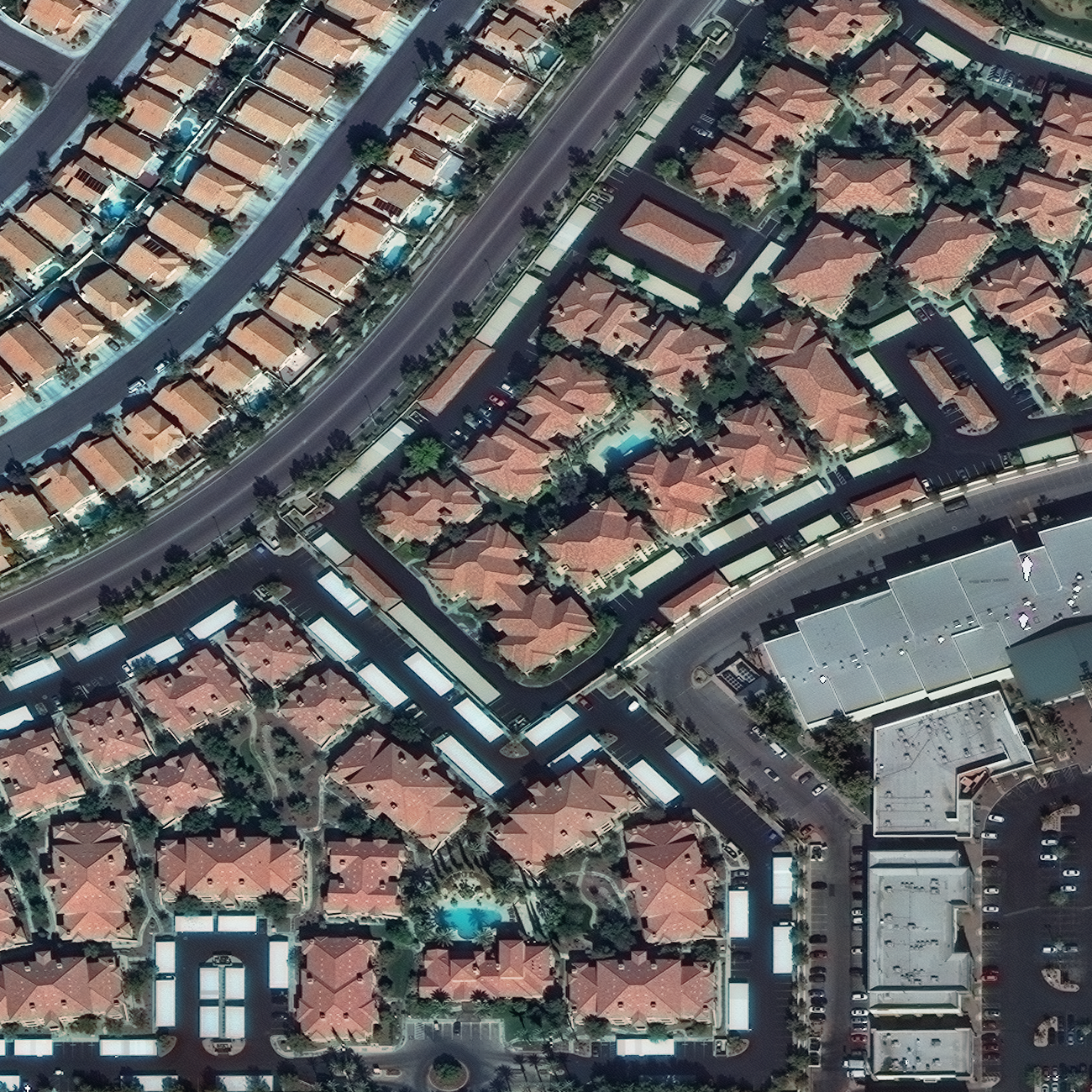} \\
		\vfil
		\includegraphics[width=1\columnwidth]{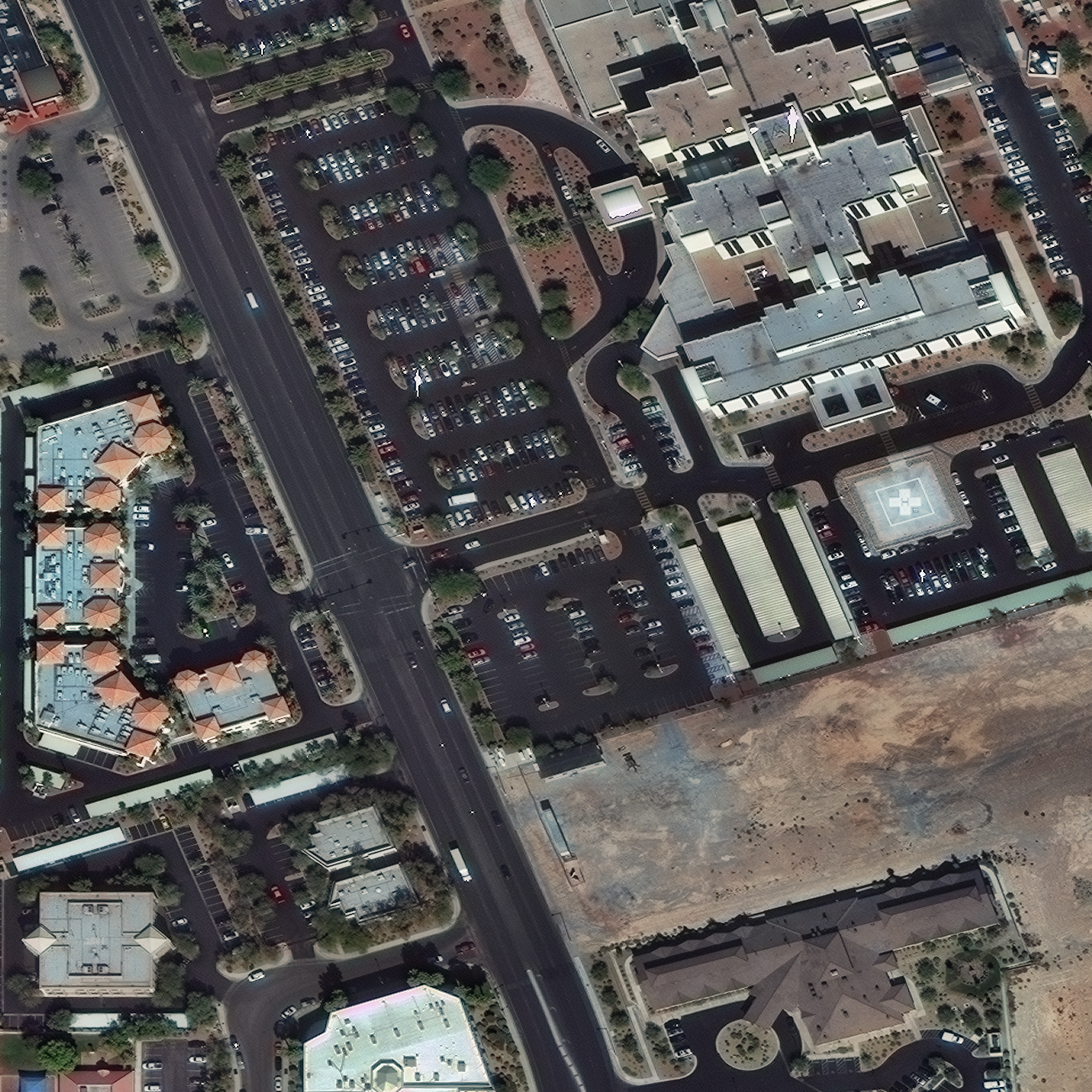} \\
	\end{minipage}
}
\subfigure[]{
	\begin{minipage}{0.2\columnwidth}
		\includegraphics[width=1\columnwidth]{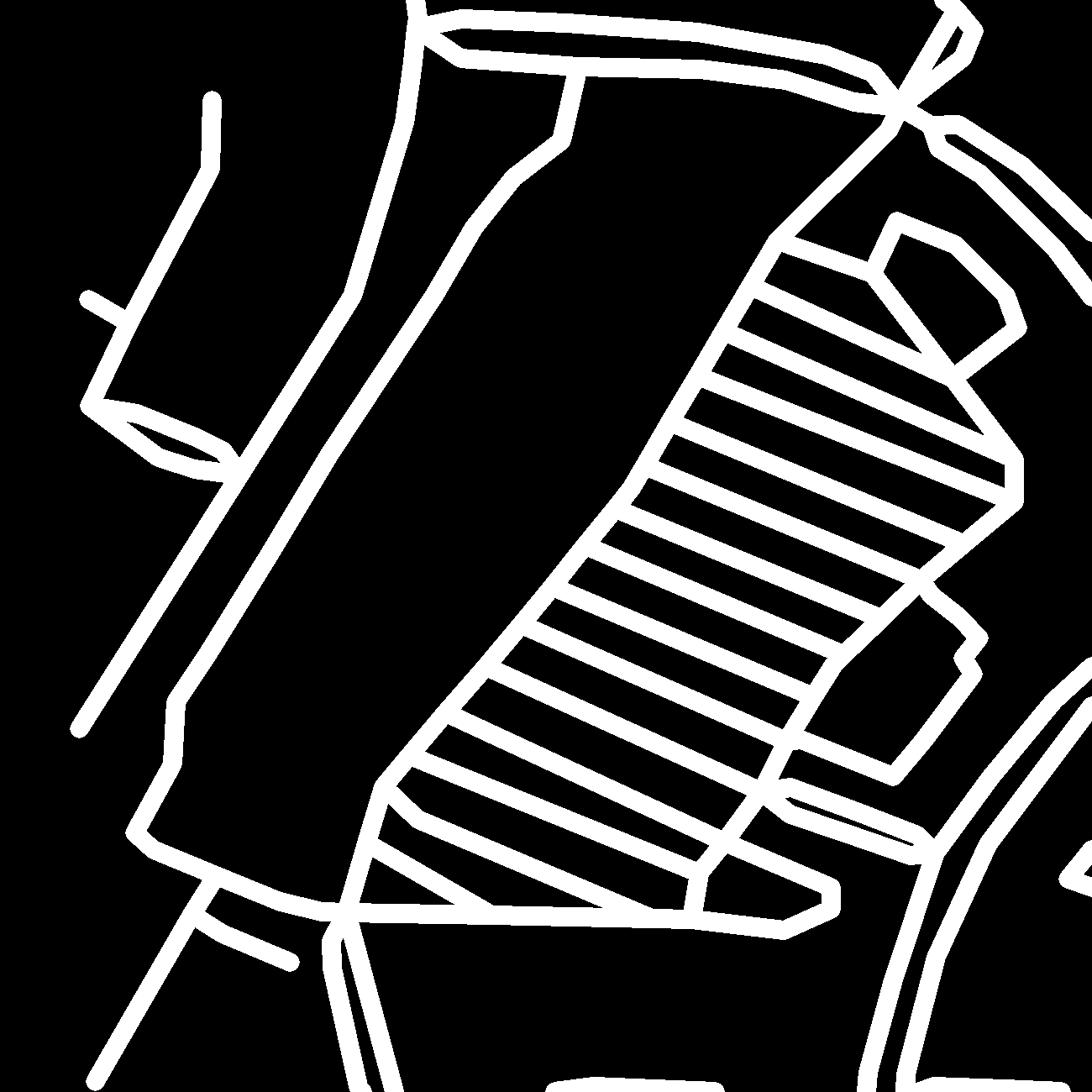} \\
		\vfil
		\includegraphics[width=1\columnwidth]{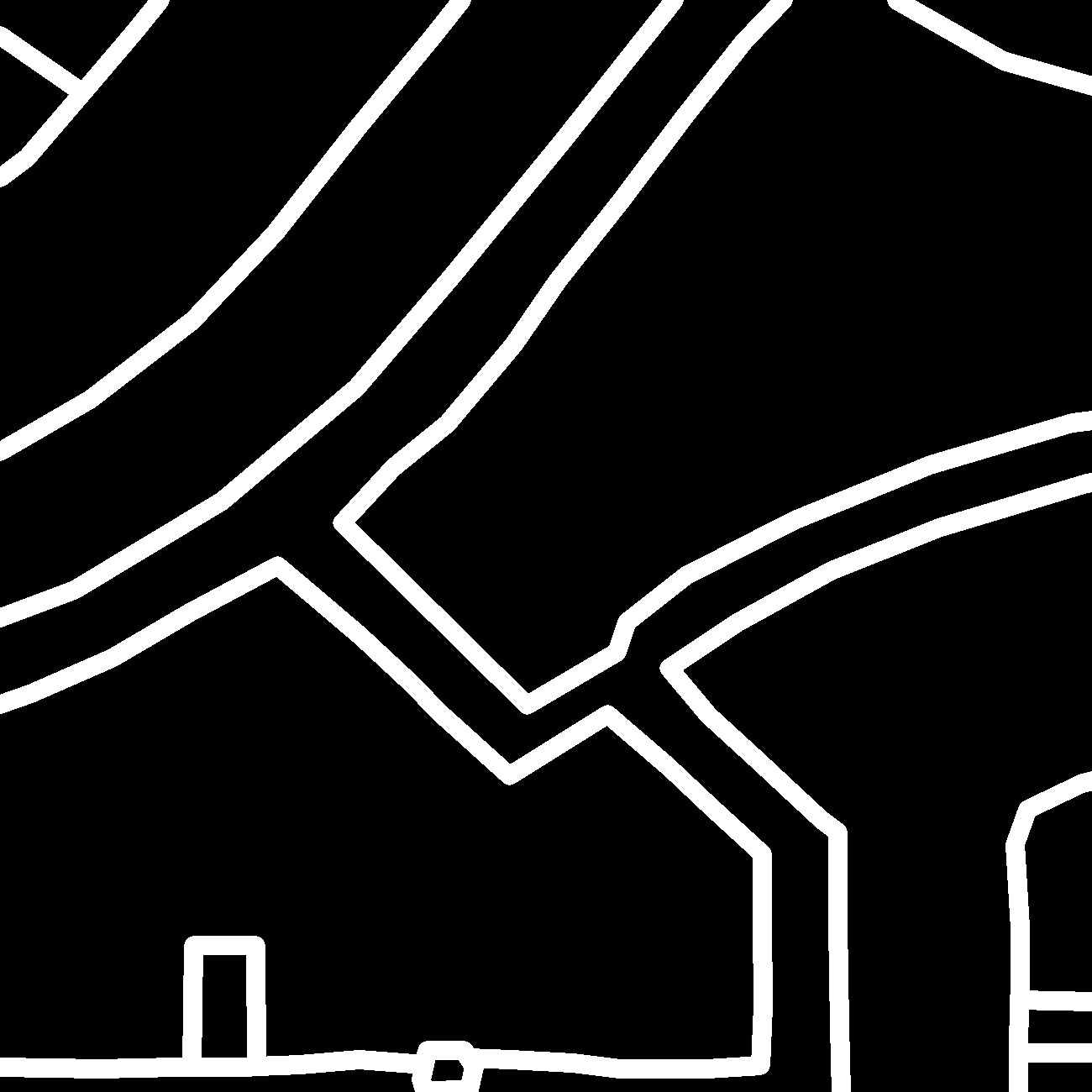} \\
		\vfil
		\includegraphics[width=1\columnwidth]{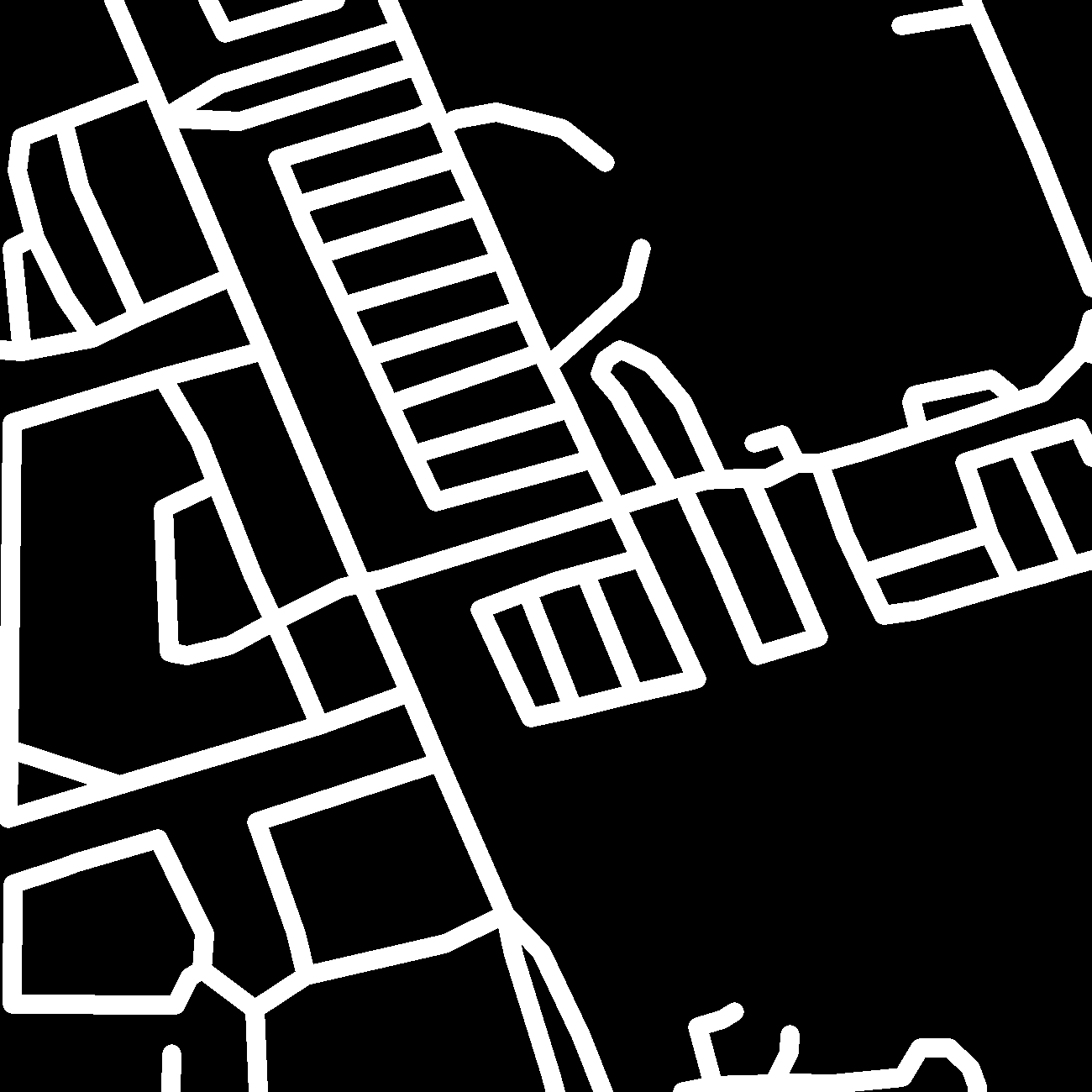} \\
	\end{minipage}
}
\subfigure[]{
	\begin{minipage}{0.2\columnwidth}
		\includegraphics[width=1\columnwidth]{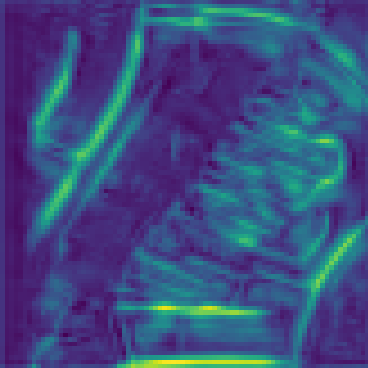} \\
		\vfil
		\includegraphics[width=1\columnwidth]{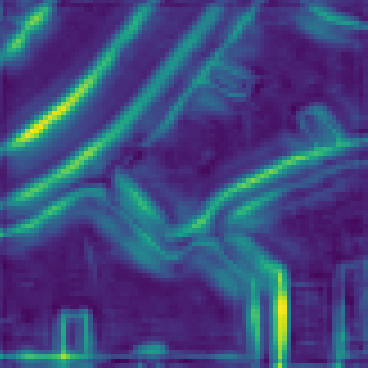} \\
		\vfil
		\includegraphics[width=1\columnwidth]{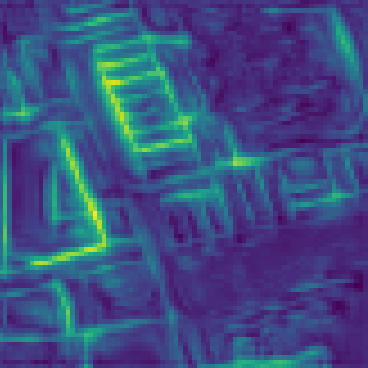} \\
	\end{minipage}
}
\subfigure[]{
	\begin{minipage}{0.2\columnwidth}
		\includegraphics[width=1\columnwidth]{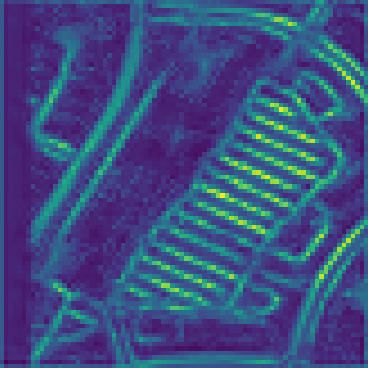} \\
		\vfil
		\includegraphics[width=1\columnwidth]{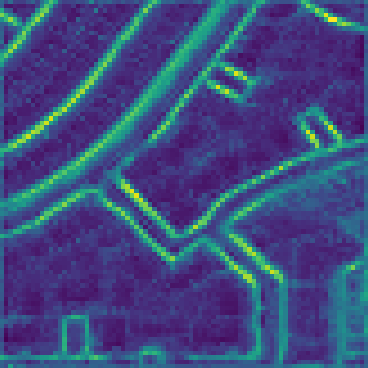} \\
		\vfil
		\includegraphics[width=1\columnwidth]{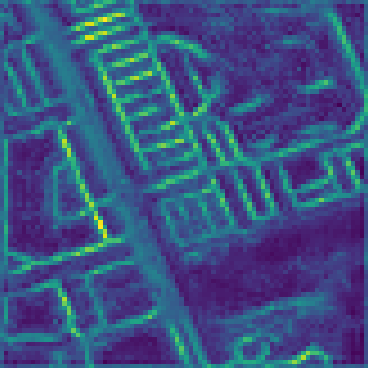} \\
	\end{minipage}
}

\caption{Visualization of feature maps near the global-aware module. (a) Original images. (b) Ground truths. (c) Feature maps from the model without GA. (d) Feature maps from models with GA. When global-aware modules are used, feature maps can more accurately express the location and details of roads.}
\label{fig:GA_com}
\end{figure}

We choose the classic semantic segmentation network, UNet, and LinkNet34, as the baselines for validating the effectiveness of our global-aware module. As listed in Table \ref{tb:GA}, the GA outperforms the baselines in three different datasets. Compared with plain UNet and LinkNet34, those with the GA module improve the IoU by 1.51\% and 1.34\% for the SpaceNet dataset, respectively. We also visualize the feature maps at the corresponding location in networks with/without the GA module in Fig. \ref{fig:GA_com}. The network with a GA module preserves more high-frequency information reflecting road details and reduces the background distraction. Thus, the GA can focus on the regions of interest and learn more accurate road features.

\begin{table}[!t]
\renewcommand{\arraystretch}{1.3}
\caption{Comparison of global attention module on road extraction results. IoU: standard intersection-over-union. APLS: average path length similarity of road graph extracted from road segmentation.}
\label{tb:GA}
\centering
\begin{tabular}{ c | c c | c c | c c }
\hline
\multirow{2}*{Methods} & 
\multicolumn{2}{c|}{\textbf{SpaceNet}} & \multicolumn{2}{c|}{\textbf{RoadTracer}} & \multicolumn{2}{c}{\textbf{Massachusetts}}\\ 
&IoU & APLS  & IoU & APLS & IoU & APLS   \\ 
\hline
LinkNet34 & 63.61 & 64.84 & 50.73 & 67.00 & 59.30 & 71.38 \\
LinkNet34+GA & \textbf{65.12} & \textbf{65.69} & \textbf{54.03} & \textbf{71.18} & 61.64 & 75.07 \\
UNet & 61.33 & 56.28 & 51.10 & 60.11 & 62.33 & 73.98 \\
UNet+GA & 62.67 & 58.26 & 52.97 & 62.16 & \textbf{62.75} & \textbf{76.84}\\
\hline

\end{tabular}
\end{table}

\subsubsection{\textbf{Connectivity Learning}} \label{sec:conn_com}

To fuse the connectivity task, we compare three combination methods in Fig. \ref{fig:MHS}: multihead network, multibranch network, and side-output network. UNet and LinkNet34 are taken as the baselines on SpaceNet. It is noted that the multihead network can achieve the best performance with the minimum parameters. Our connectivity task improves the APLS by 2.98\% and 2.28\% compared with the two baselines. The multihead network is more stable in obtaining outputs without the checkerboard effect caused by the transconvolution layers.

\begin{table}[!t]
\renewcommand{\arraystretch}{1.3}
\caption{Comprising performance of the 3 multitask architectures, where MH, MB, SO represent Multihead, Multibranch and Side-Output, separately. (Due to the GPU memory limitation, UNet only adjusts to multihead and multibranch structures, which demonstrates that the side-output architecture is poor in inference performance). Layers: number of convolution layers. Params: the size of the model parameters. Infer Time: inference time of each $512\times 512$ image patch. IoU and APLS: metrics.}
\label{tb:MTS}
\centering
\begin{tabular}{ c | c | c | c | c | c }
\hline
Methods & Layers & \makecell[c]{Params \\ (M)} & \makecell[c]{Infer Time \\ (ms)} & IoU & APLS \\
\hline
LinkNet & 51 & 21.64 & 207 & 63.61 & 64.84 \\
LinkNet+MH & 54 & 21.67 & 225 & \textbf{66.14} & \textbf{67.82} \\
LinkNet+MB & 66 & 22.00 & 264 & 66.02 & 67.75 \\
LinkNet+SO & 96 & 26.51 & 313 & 66.07 & 67.60 \\
UNet & 18 & 7.70 & 4565 & 61.33 & 56.28 \\
UNet+MH & 21 & 7.81 & 5701 & 61.92 & 58.46 \\
UNet+MB & 28 & 10.71 & 7137 & 62.20 & 58.34 \\
\hline

\end{tabular}
\end{table}

\begin{table}[!t]
	\renewcommand{\arraystretch}{1.3}
	\caption{Comparisons of connectivity (conn), orientation (orient), and point-centerline (p-l) learning as auxiliary tasks. This shows that improved road connectivity is due to connectivity tasks rather than multitask learning.}
	\label{tb:MT}
	\centering
	\begin{tabular}{ c | c c | c c | c c }
		\hline
		\multirow{2}*{Methods} & 
		\multicolumn{2}{c|}{\textbf{SpaceNet}} & \multicolumn{2}{c|}{\textbf{RoadTracer}} & \multicolumn{2}{c}{\textbf{Massachusetts}}\\ 
		&IoU & APLS  & IoU & APLS & IoU & APLS   \\ 
		\hline
		LinkNet & 63.61 & 64.84 & 50.73 & 67.00 & 59.30 & 71.38 \\
		LinkNet+p-l & 64.50 & 65.81 & 53.96 & 72.55 & 62.07 & 75.94 \\
		LinkNet+orient & 65.11 & 65.77 & 51.93 & 62.41 & 61.00 & 74.95 \\
		LinkNet+conn & \textbf{66.14} & \textbf{67.82} & \textbf{54.92} & \textbf{74.49} & \textbf{62.57} & \textbf{76.60} \\
		\hline
		
	\end{tabular}
\end{table}

\begin{figure*}[!t]
	\centering
	\subfigure[]{
		\begin{minipage}{0.2\columnwidth}
			\includegraphics[width=1\columnwidth]{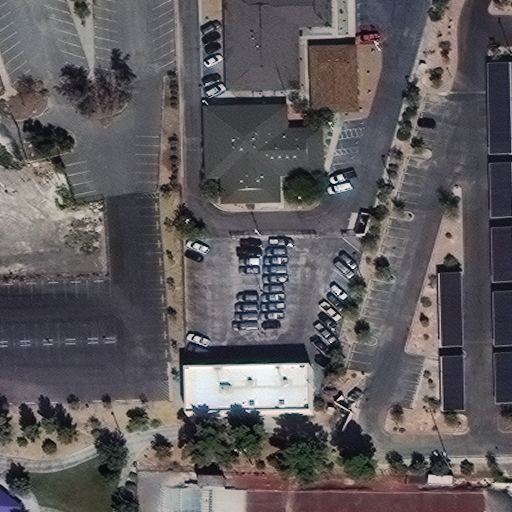} \\
			\vfill
			\includegraphics[width=1\columnwidth]{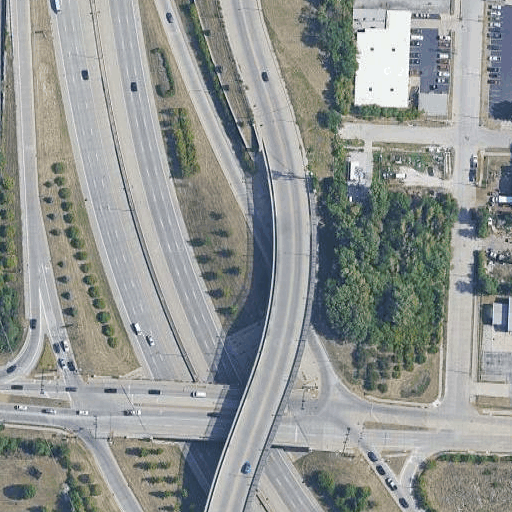} \\
			\vfil
			\includegraphics[width=1\columnwidth]{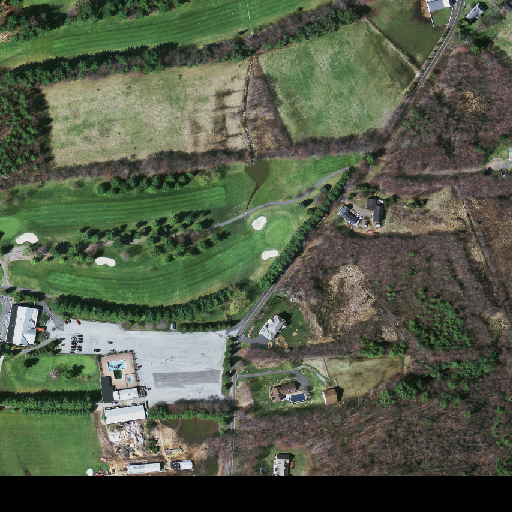} \\
		\end{minipage}
	}
	\subfigure[]{
		\begin{minipage}{0.2\columnwidth}
			\includegraphics[width=1\columnwidth]{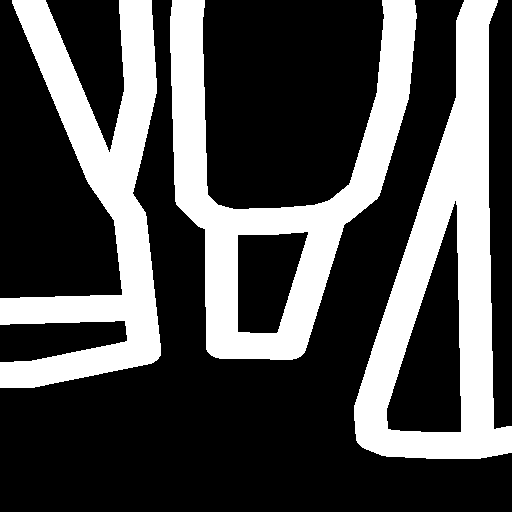} \\
			\vfil
			\includegraphics[width=1\columnwidth]{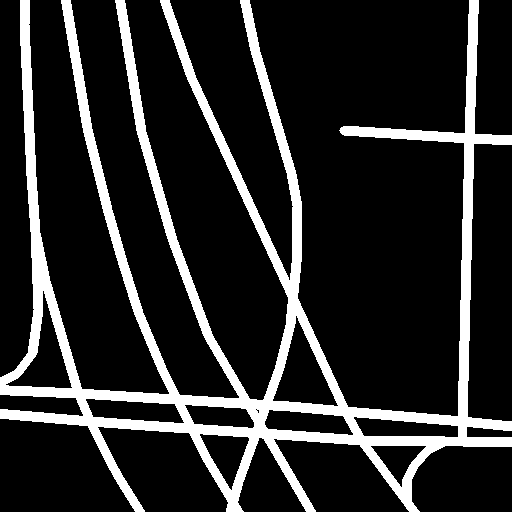} \\
			\vfil
			\includegraphics[width=1\columnwidth]{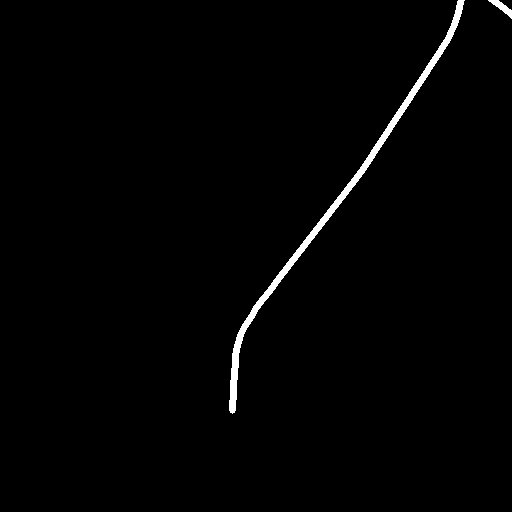} \\
		\end{minipage}
	}
	\subfigure[]{
		\begin{minipage}{0.2\columnwidth}
			\includegraphics[width=1\columnwidth]{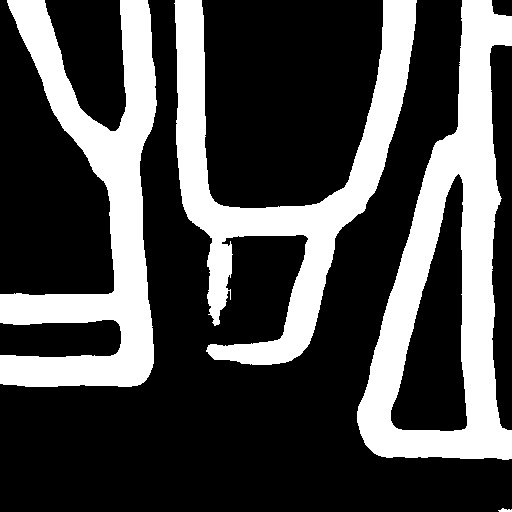} \\
			\vfil
			\includegraphics[width=1\columnwidth]{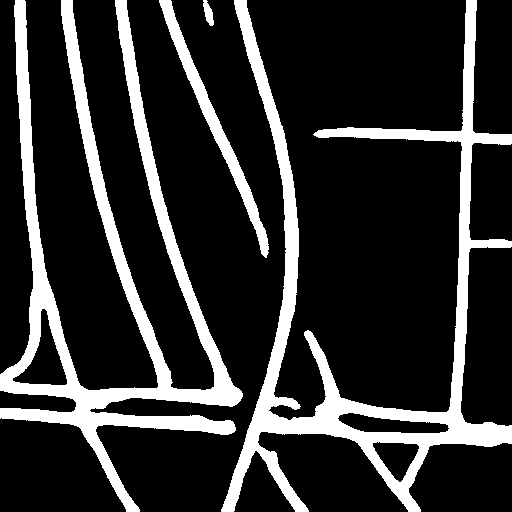} \\
			\vfil
			\includegraphics[width=1\columnwidth]{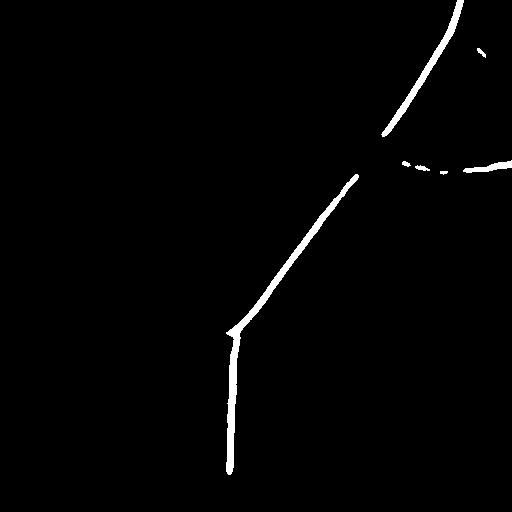} \\
		\end{minipage}
	}
	\subfigure[]{
		\begin{minipage}{0.2\columnwidth}
			\includegraphics[width=1\columnwidth]{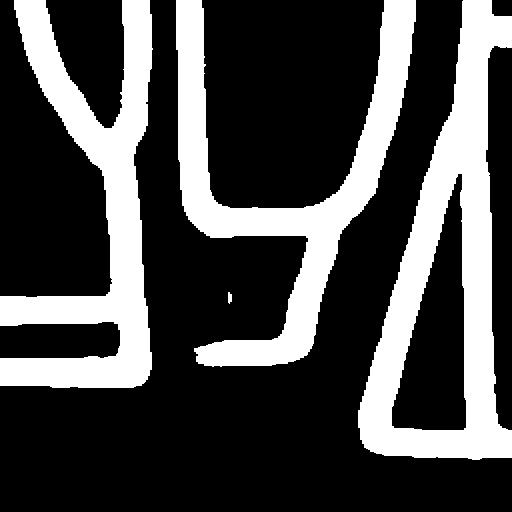} \\
			\vfil
			\includegraphics[width=1\columnwidth]{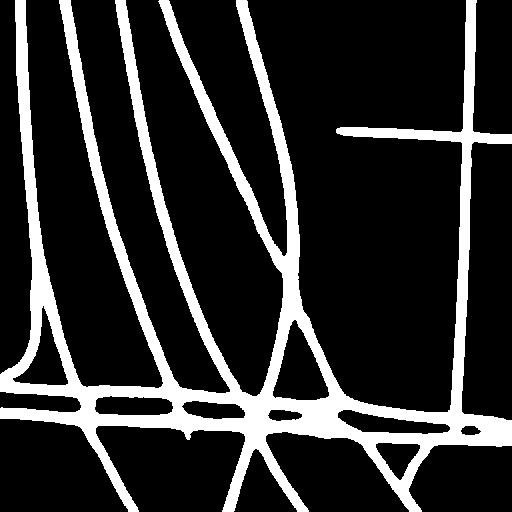} \\
			\vfil
			\includegraphics[width=1\columnwidth]{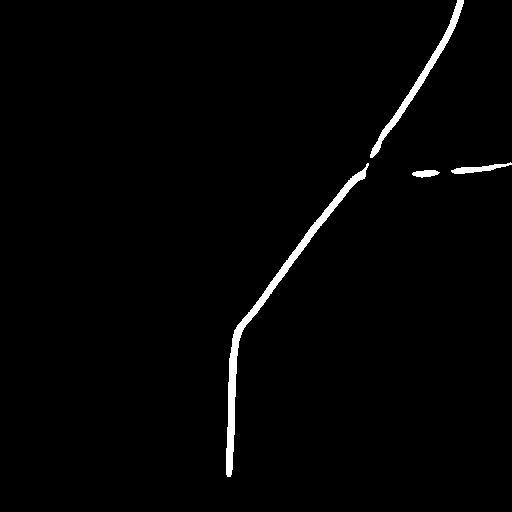} \\
		\end{minipage}
	}
	\subfigure[]{
		\begin{minipage}{0.2\columnwidth}
			\includegraphics[width=1\columnwidth]{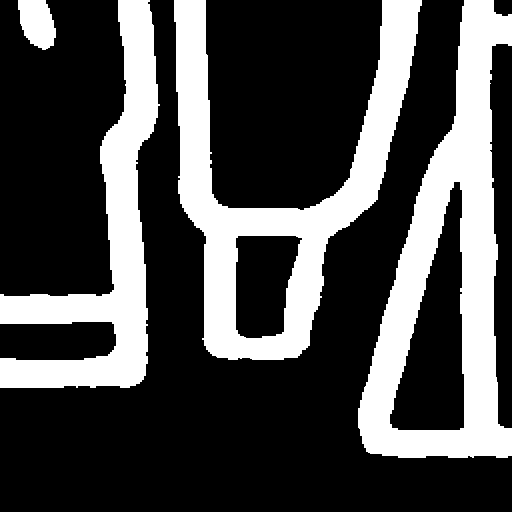} \\
			\vfil
			\includegraphics[width=1\columnwidth]{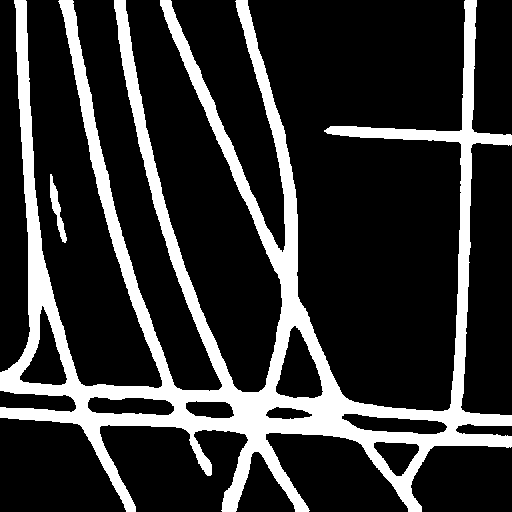} \\
			\vfil
			\includegraphics[width=1\columnwidth]{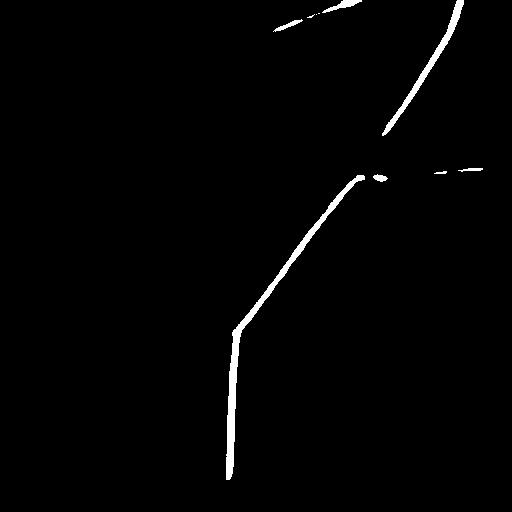} \\
		\end{minipage}
	}
	\subfigure[]{
		\begin{minipage}{0.2\columnwidth}
			\includegraphics[width=1\columnwidth]{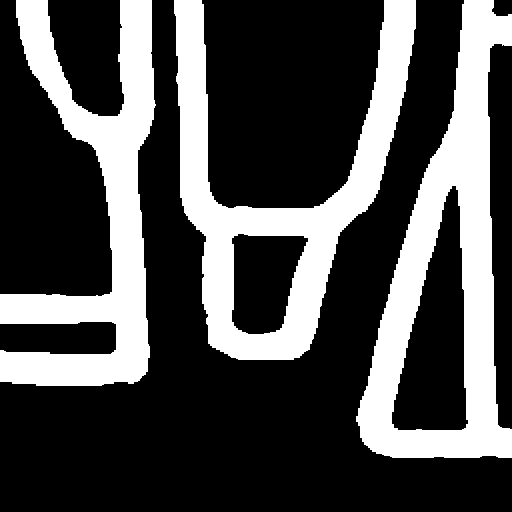} \\
			\vfil
			\includegraphics[width=1\columnwidth]{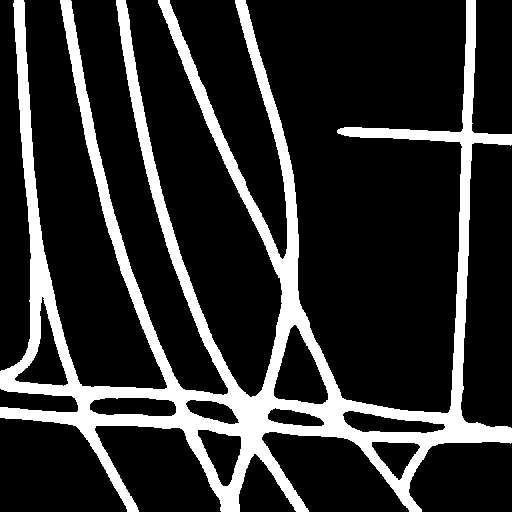} \\
			\vfil
			\includegraphics[width=1\columnwidth]{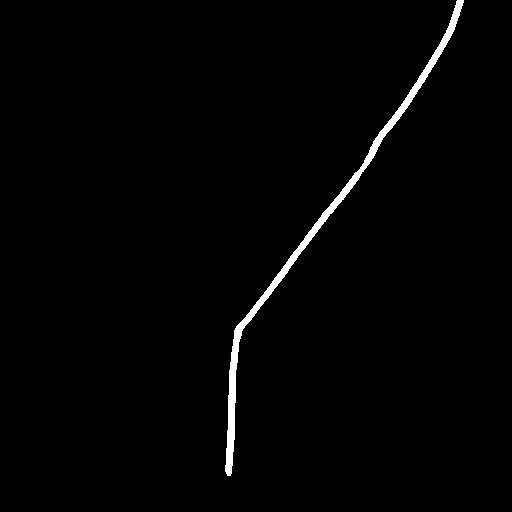} \\
		\end{minipage}
	}
	\caption{Predictions of LinkNet34 with different auxiliary tasks. (a) original images. (b) Ground-truth images (c) Plain LinkNet34 predictions. (d) LinkNet with point-line task predictions. (e) LinkNet with orientation task predictions. (f) LinkNet with our connectivity task predictions.}
	\label{fig:Conn-com}
\end{figure*}

To study the significance of the connectivity task for preserving the correct topological relationships among the road segments, we adjusted LinkNet34 into the multihead structure and compared it with the other two types of shared tasks. The first is the orientation task used in \cite{batra_improved_2019}. The second is a point-line task representing road centerlines and intersections, which is considered because points and lines are components of road graphs, so intuitively, jointly learning these two theoretically improves road connectivity. The results listed in Table \ref{tb:MT} show that the APLS is improved with the connectivity task and is higher than those with the orientation task or point-line task, indicating that the connectivity task in our method is effective for predicting the topology features of the connected roads. This suggests that the improvement in the results is due to our connectivity task rather than to the multitask learning mechanism. Fig. \ref{fig:Conn-com} shows the segmentation outputs of the above networks in three different scenes. We discover that incorporating our connectivity task performs better in road connectivity in parking lot lanes, roads under the overpass, and unpaved tracks shaded by trees than other networks.

\subsubsection{\textbf{Stacked Multihead Network}} \label{sec:smn}

To verify the superiority of the stacked structure over the commonly used plain structure, we compare our stacked multihead network with the state-of-the-art semantic segmentation network. UNet and LinkNet are adjusted into multihead forms, and we also study the impact of the stack number on the performance. The results in Table \ref{tb:nstack} show that the multihead network with two stacks is the most stable, and the IoU and APLS increase by an average of 4\% and 8\%, respectively, over the UNet multihead structure.

\begin{table}[!t]
\renewcommand{\arraystretch}{1.3}
\caption{Comparison of our stacked network and commonly used segmentation networks adjusted into multihead architectures for joint learning and comparing the effect of the number of stacks. It shows that the 2 stacked architectures improve the IoU and APLS by approximately 4\% and 8\%, respectively.}
\label{tb:nstack}
\centering
\begin{tabular}{ c | c c | c c | c c }
\hline
\multirow{2}*{Methods} & 
\multicolumn{2}{c|}{\textbf{SpaceNet}} & \multicolumn{2}{c|}{\textbf{RoadTracer}} & \multicolumn{2}{c}{\textbf{Massachusetts}}\\ 
&IoU & APLS  & IoU & APLS & IoU & APLS   \\ 
\hline
UNet+connectivity & 61.92 & 58.46 & 51.93 & 62.41 & 62.71 & 74.81 \\
LinkNet+conn & 66.14 & 67.82 & 54.92 & 74.49 & 62.57 & 76.60 \\
1 Stack+conn & 66.55 & 67.88 & 56.22 & 75.37 & 61.24 & 74.29 \\
2 Stacks+conn & \textbf{67.08} & \textbf{68.36} & \textbf{56.33} & \textbf{75.76} & \textbf{63.50} & \textbf{77.75} \\
3 Stacks+conn & 66.80 & 68.30 & 56.28 & 75.46 & 63.05 & 77.35 \\
\hline
\end{tabular}
\end{table}

To demonstrate the effectiveness of our proposed GA module and connectivity task, we incrementally apply them to our framework. As shown in Table \ref{tb:inc}, we set the network with two stacks as the baseline and then add the global-aware module and the road connectivity task. The final form of our network is a 2-stack 2-head network with GA modules for simultaneously learning road segmentation and road connectivity. Our final network performs better on almost all datasets, indicating that the global awareness and connectivity tasks both contribute to improving road extraction from remote sensing images. However, it can be seen that the global-aware module has a slight improvement on this stacked network compared with LinkNet and UNet. We hypothesize that 1/4 intermediate supervision works in a similar way to the global attention mechanism.

\begin{table}[!t]
\renewcommand{\arraystretch}{1.3}
\caption{The incremental improvement of our proposed methods (global-aware module and road connectivity task).}
\label{tb:inc}
\centering
\begin{tabular}{ c | c c | c c | c c }
\hline
\multirow{2}*{Methods} & 
\multicolumn{2}{c|}{\textbf{SpaceNet}} & \multicolumn{2}{c|}{\textbf{RoadTracer}} & \multicolumn{2}{c}{\textbf{Massachusetts}}\\ 
&IoU & APLS  & IoU & APLS & IoU & APLS   \\ 
\hline
2 Stacks & 65.86 & 66.82 & 54.88 & 73.24 & 62.56 & 76.78 \\
2 Stacks+GA & 66.04 & 67.40 & 55.31 & 73.43 & 62.88 & 76.81 \\
2 Stacks+conn & 67.08 & 68.36 & 56.33 & 75.76 & \textbf{63.50} & 77.74 \\
2 Stacks+GA+conn & \textbf{67.51} & \textbf{68.87} & \textbf{56.63} & \textbf{76.18} & 63.38 & \textbf{77.75} \\
\hline
\end{tabular}
\end{table}

\begin{figure*}[!t]
	\centering
	\includegraphics[width=\textwidth]{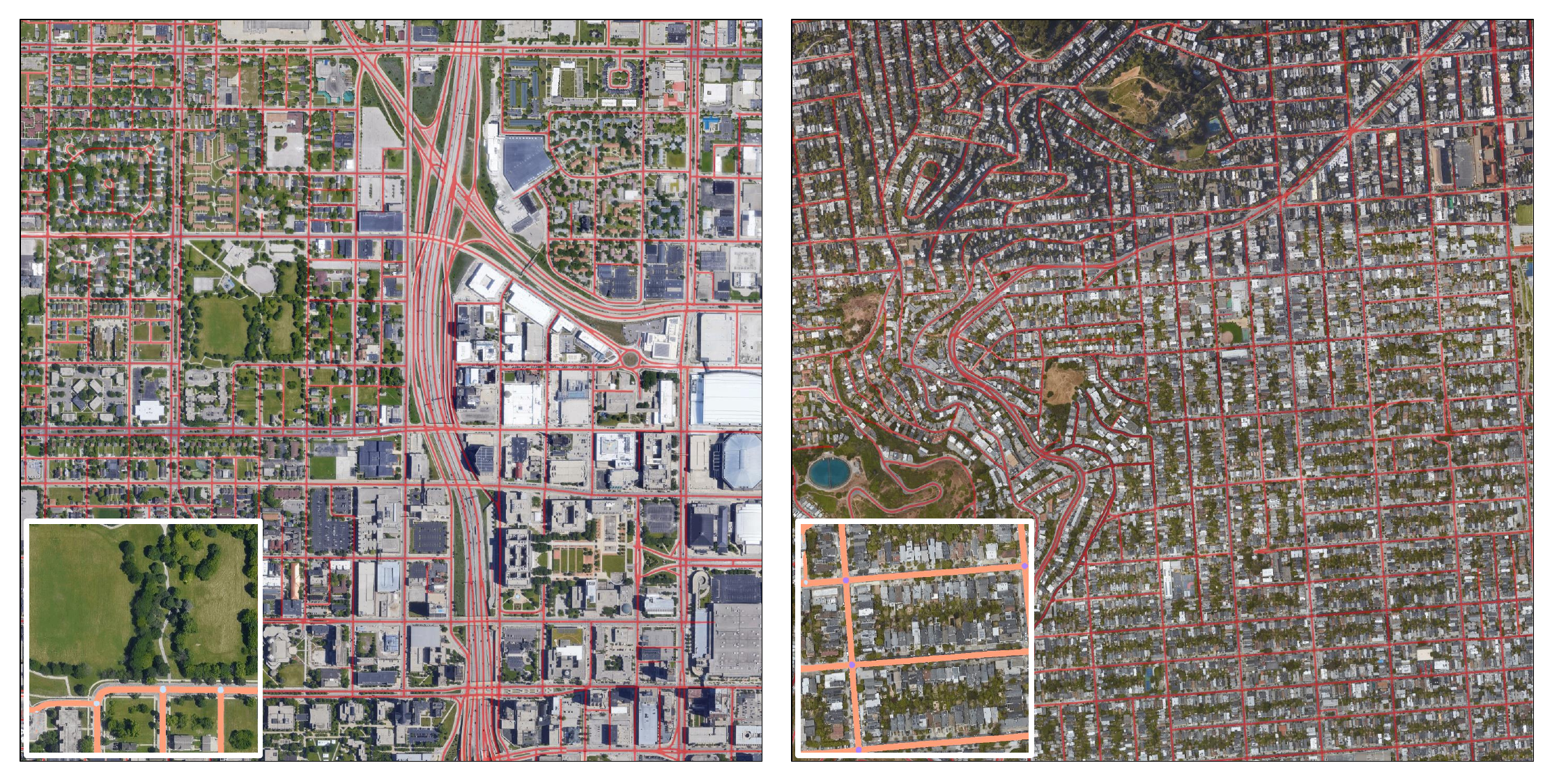}
	\caption{Example prediction of two large-scale remote sensing images. The red areas in each image indicate roads extracted by our method. In the zoomed-in view at the bottom left of each image, the prediction of connectivity is displayed, where orange areas represent the road, and blue and purple areas represent intersections with 3 and 4 branches, respectively.}
	\label{fig:LS}
\end{figure*}

To demonstrate the effectiveness of our method over large-scale remote sensing images, we conduct inference experiments on the original RoadTracer test imagery ($4,096\times 4,096$). Since the input size of the network is confined to the GPU memory size and the inference images are relatively large, we need to crop and splice the images in the inference process. Because of the lack of context information in the edge area of each patch, the boundary prediction is not that accurate. To avoid splicing marks caused by the boundary effect, we adopt expansion prediction. We set the inference patch size as $512\times 512$, the step size as 368, ignore the 72 pixels at the edge and splice into the final prediction. The results of two $4,096\times 4,096$ images are shown in Fig. \ref{fig:LS}. These two images are from a different country with different landscapes and contain a hybrid of roads with various scales and appearances, which makes road extraction very challenging. Our network can identify most of the roads, which proves its effectiveness and universality. The zoomed-in view at the bottom left of each image displays the connectivity result. Our network can accurately count the branches of each intersection and identify road areas, which allows the network to learn the topological relationship of the road segments and improve the connectivity of the road graph, especially at the intersection.

\subsubsection{\textbf{Comparisons with State-of-the-Art Methods}}

We compare our proposed methods with the state-of-the-art segmentation or road segmentation methods UNet \cite{ronneberger_u-net:_2015}, LinkNet \cite{chaurasia--linknet_2017}, DLA \cite{yu_deep_2019}, D-LinkNet \cite{zhou_d-linknet:_2018}, DeepRoadMapper (segmentation) \cite{mattyus_deeproadmapper_2017}, Li \textit{et al.} \cite{batra_improved_2019
} and Batra \textit{et al.} \cite{li_topology-enhanced_2020}. The performances on the three datasets are listed in Table \ref{tb:comparison}, and from this table, we observe that our method outperforms the others.

\begin{figure*}[!t]
	\centering
	\setcounter{subfigure}{0}
	\begin{minipage}{0.01\textwidth}
		\footnotesize
		\vfill
		(a) \\
		\vspace{1.05cm}
		(b) \\
		\vspace{1.05cm}
		(c) \\
		\vspace{1.05cm}
		(d) \\
		\vspace{1.05cm}
		(e) \\
		\vspace{1.05cm}
		(f)
		\vfill
	\end{minipage}
	\subfigure{
		\begin{minipage}{0.07\textwidth}
			\includegraphics[width=1.23\columnwidth]{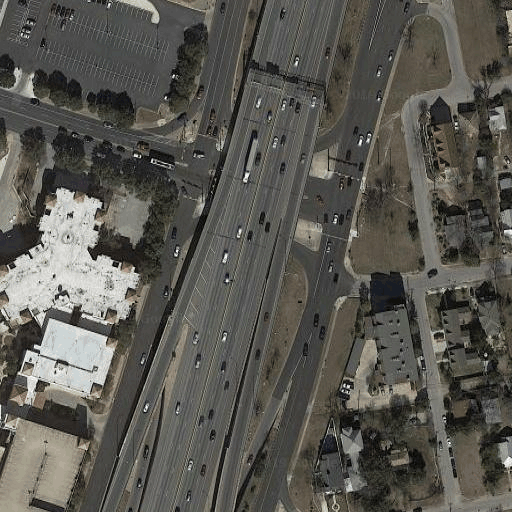} \\
			\vspace{-0.3cm}
			\includegraphics[width=1.23\columnwidth]{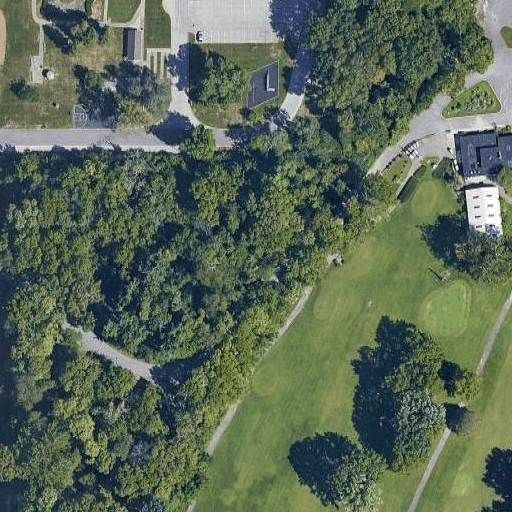} \\
			\vspace{-0.3cm}
			\includegraphics[width=1.23\columnwidth]{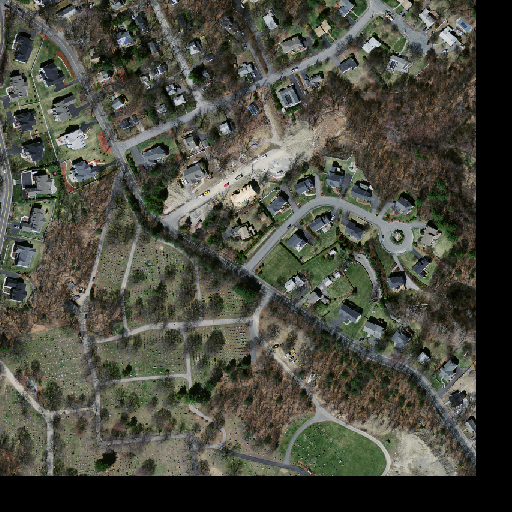} \\
			\vspace{-0.3cm}
			\includegraphics[width=1.23\columnwidth]{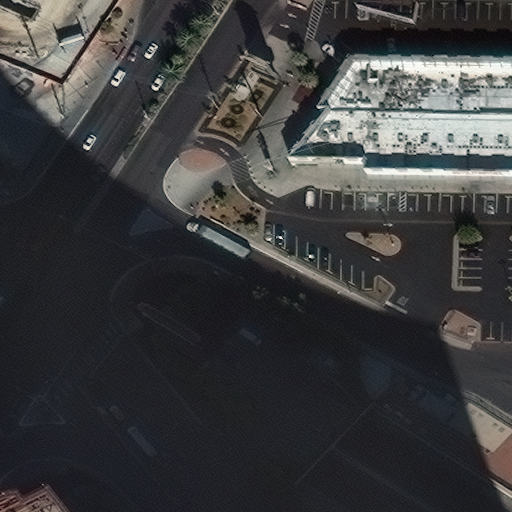} \\
			\vspace{-0.3cm}
			\includegraphics[width=1.23\columnwidth]{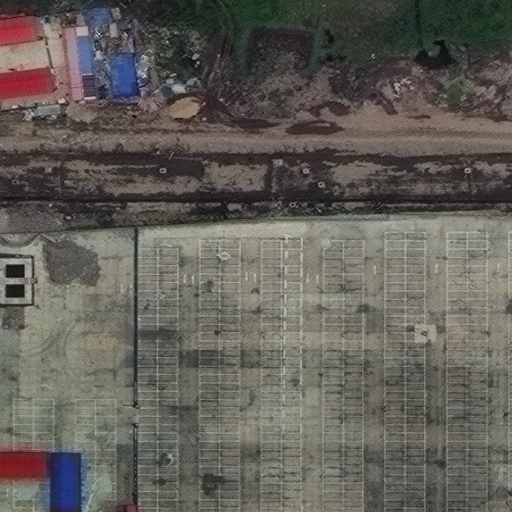} \\
			\vspace{-0.3cm}
			\includegraphics[width=1.23\columnwidth]{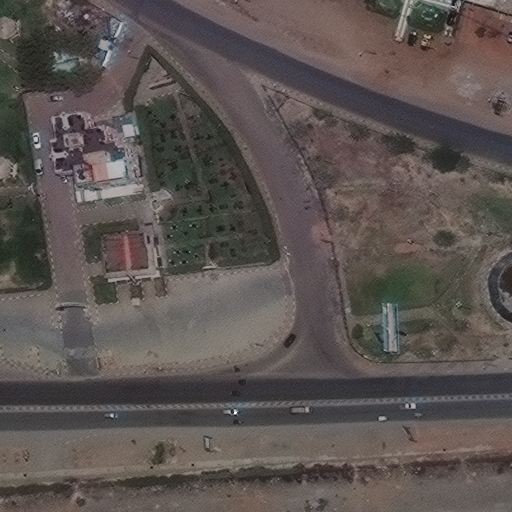}
		\end{minipage}
	}
	\subfigure{
		\begin{minipage}{0.07\textwidth}
			\includegraphics[width=1.23\columnwidth]{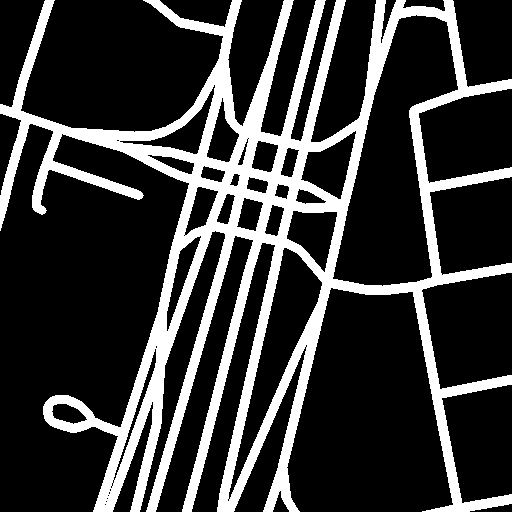} \\
			\vspace{-0.3cm}
			\includegraphics[width=1.23\columnwidth]{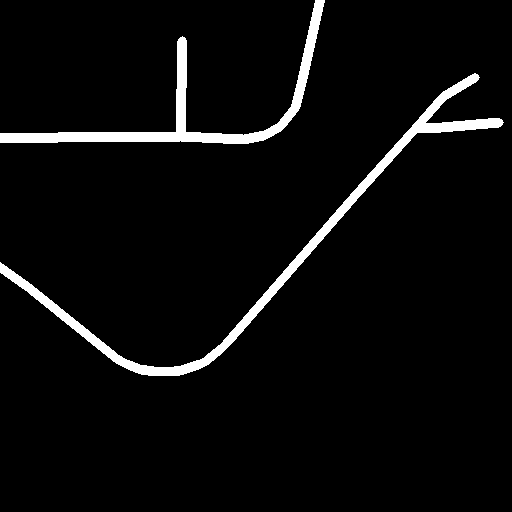} \\
			\vspace{-0.3cm}
			\includegraphics[width=1.23\columnwidth]{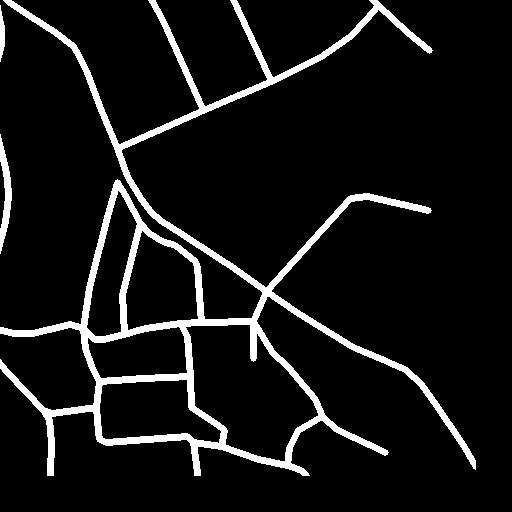} \\
			\vspace{-0.3cm}
			\includegraphics[width=1.23\columnwidth]{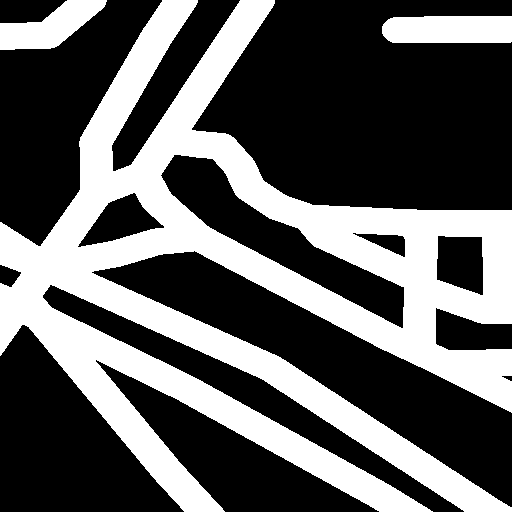} \\
			\vspace{-0.3cm}
			\includegraphics[width=1.23\columnwidth]{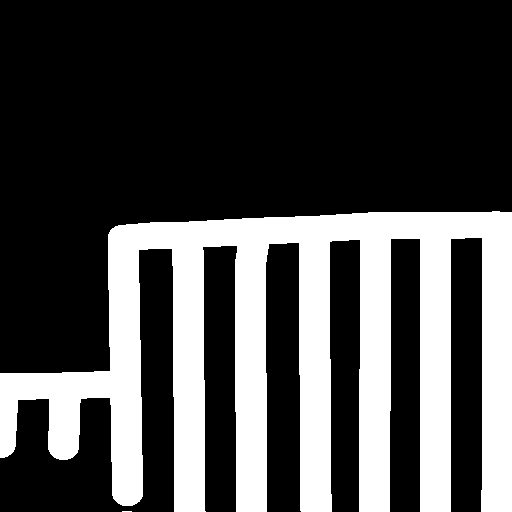} \\
			\vspace{-0.3cm}
			\includegraphics[width=1.23\columnwidth]{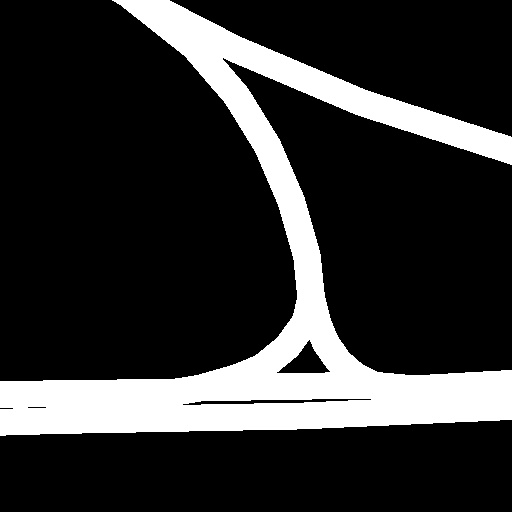}
		\end{minipage}
	}
	\subfigure{
		\begin{minipage}{0.07\textwidth}
			\includegraphics[width=1.23\columnwidth]{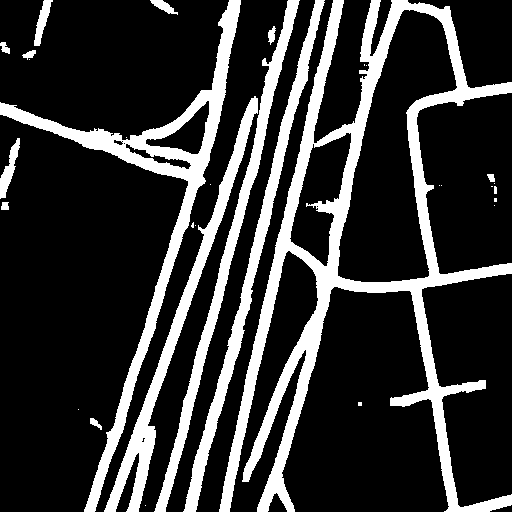} \\
			\vspace{-0.3cm}
			\includegraphics[width=1.23\columnwidth]{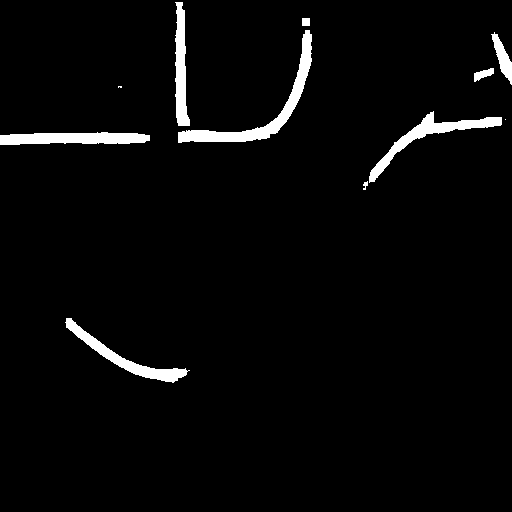} \\
			\vspace{-0.3cm}
			\includegraphics[width=1.23\columnwidth]{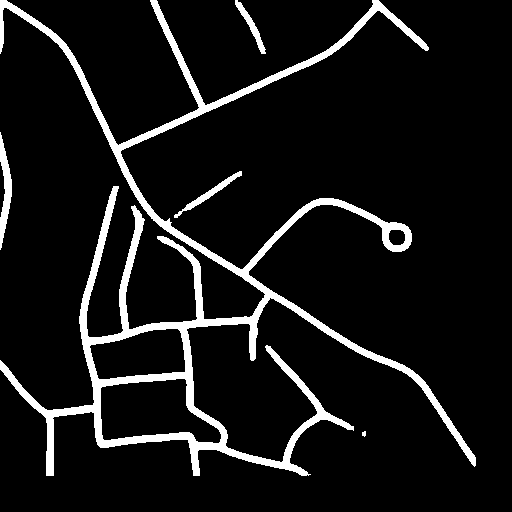} \\
			\vspace{-0.3cm}
			\includegraphics[width=1.23\columnwidth]{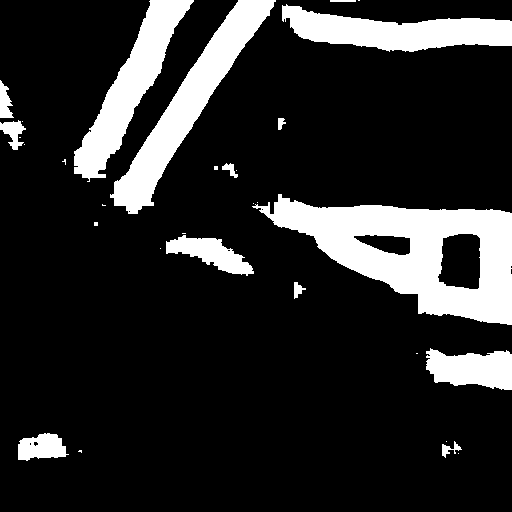} \\
			\vspace{-0.3cm}
			\includegraphics[width=1.23\columnwidth]{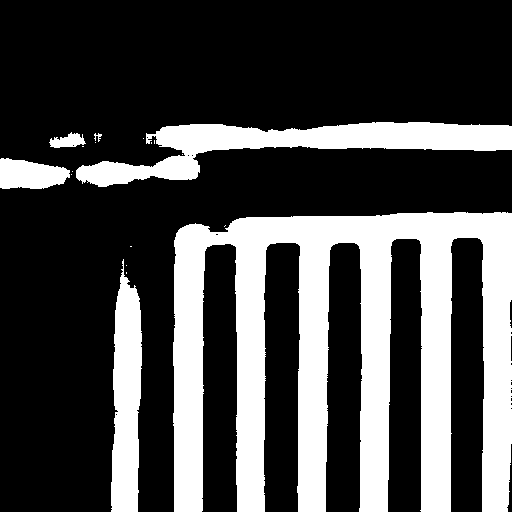} \\
			\vspace{-0.3cm}
			\includegraphics[width=1.23\columnwidth]{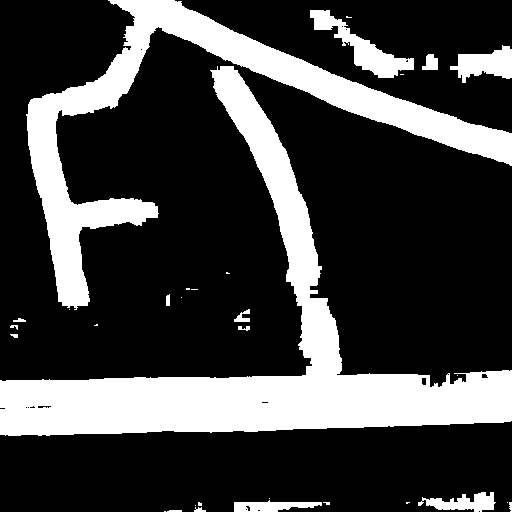}
		\end{minipage}
	}
	\subfigure{
		\begin{minipage}{0.07\textwidth}
			\includegraphics[width=1.23\columnwidth]{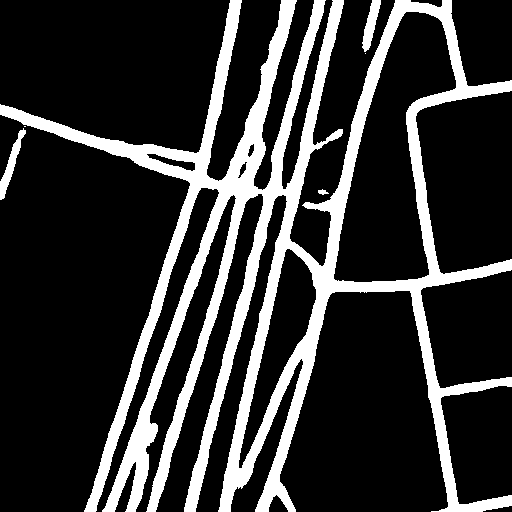} \\
			\vspace{-0.3cm}
			\includegraphics[width=1.23\columnwidth]{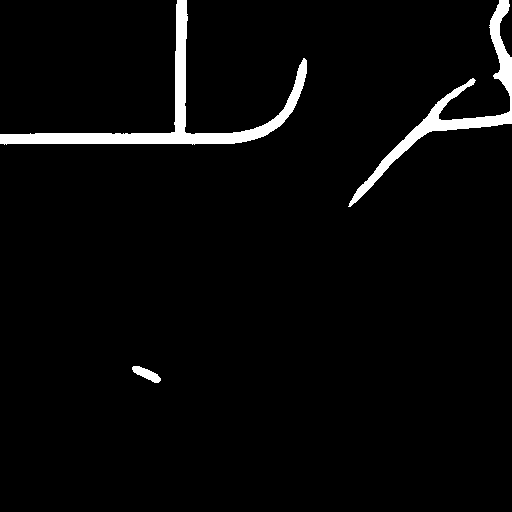} \\
			\vspace{-0.3cm}
			\includegraphics[width=1.23\columnwidth]{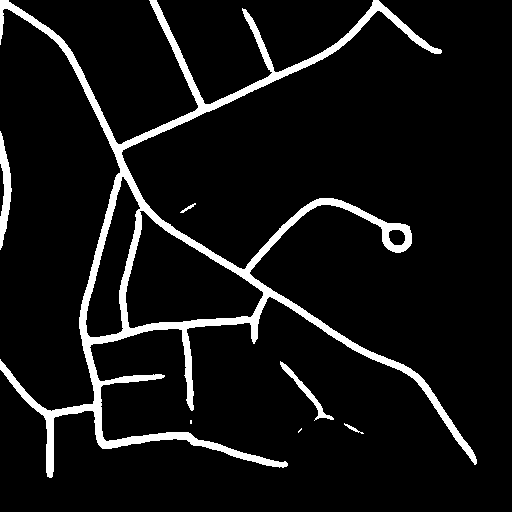} \\
			\vspace{-0.3cm}
			\includegraphics[width=1.23\columnwidth]{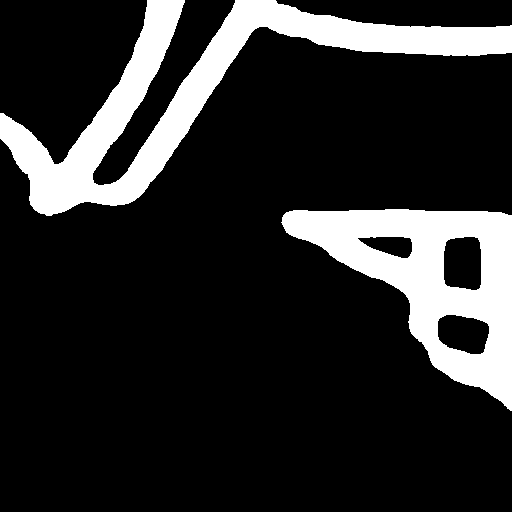} \\
			\vspace{-0.3cm}
			\includegraphics[width=1.23\columnwidth]{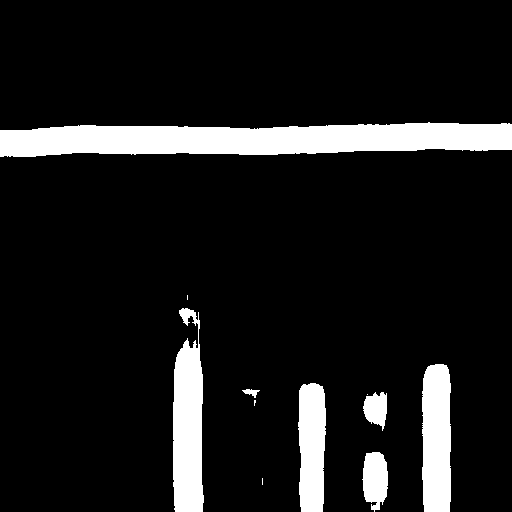} \\
			\vspace{-0.3cm}
			\includegraphics[width=1.23\columnwidth]{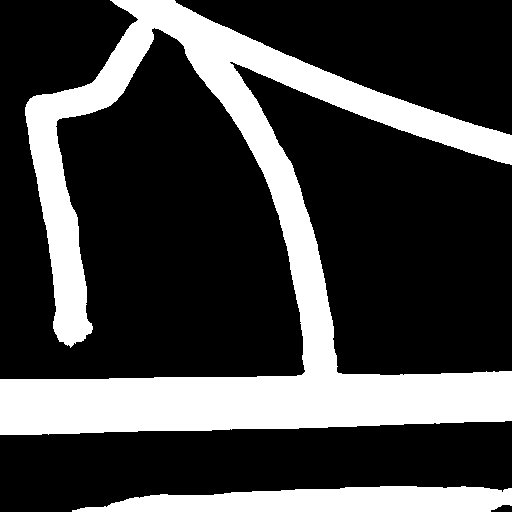}
		\end{minipage}
	}
	\subfigure{
		\begin{minipage}{0.07\textwidth}
			\includegraphics[width=1.23\columnwidth]{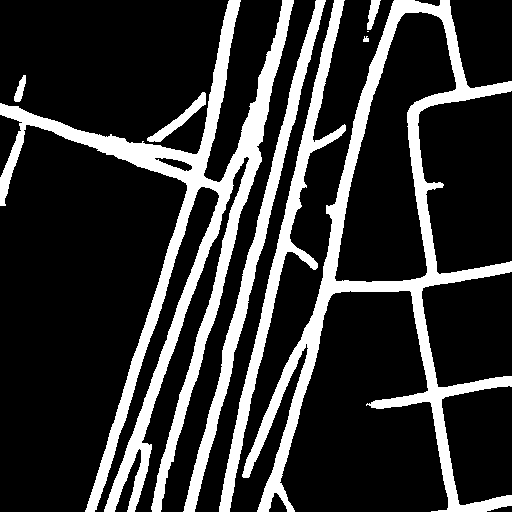} \\
			\vspace{-0.3cm}
			\includegraphics[width=1.23\columnwidth]{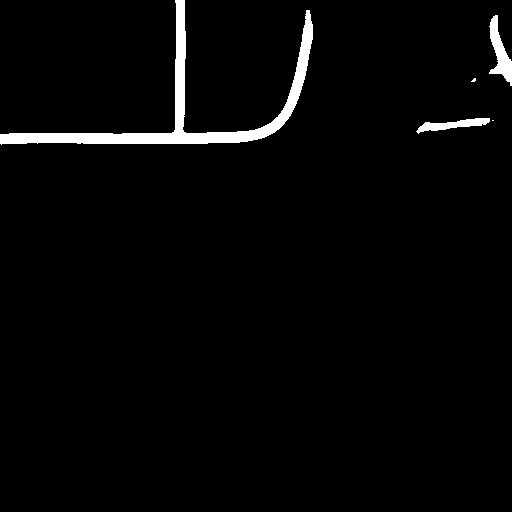} \\
			\vspace{-0.3cm}
			\includegraphics[width=1.23\columnwidth]{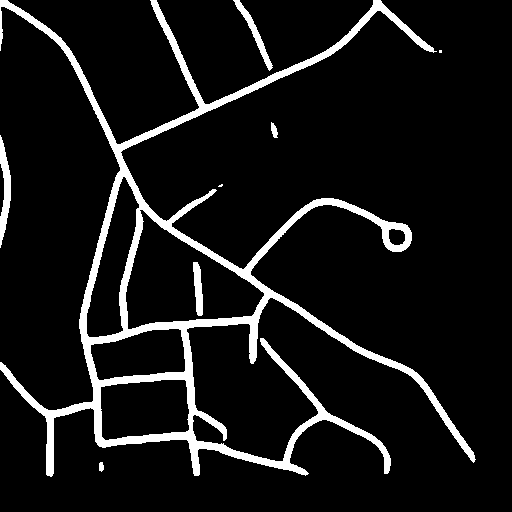} \\
			\vspace{-0.3cm}
			\includegraphics[width=1.23\columnwidth]{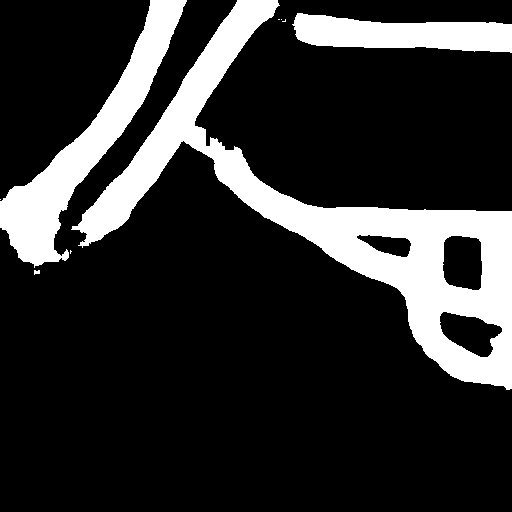} \\
			\vspace{-0.3cm}
			\includegraphics[width=1.23\columnwidth]{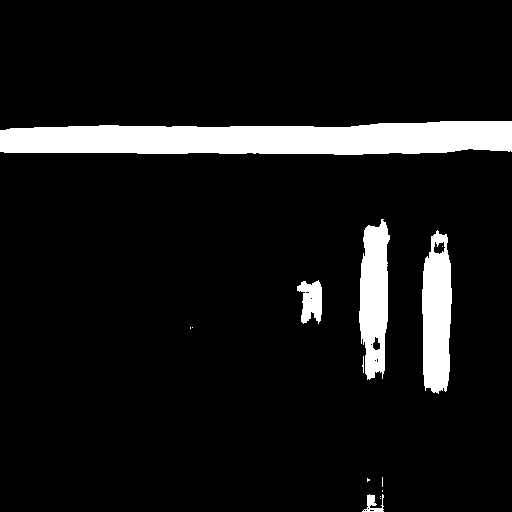} \\
			\vspace{-0.3cm}
			\includegraphics[width=1.23\columnwidth]{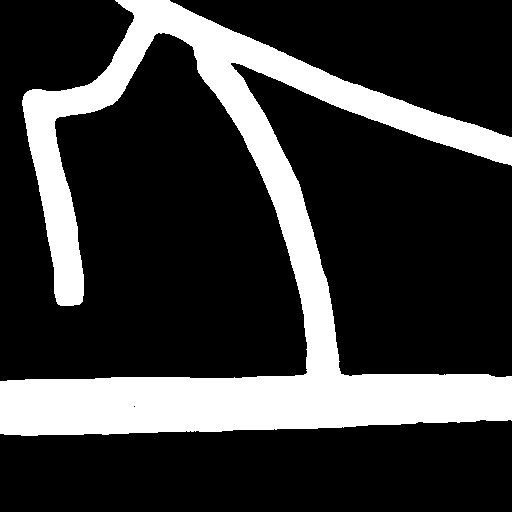}
		\end{minipage}
	}
	\subfigure{
		\begin{minipage}{0.07\textwidth}
			\includegraphics[width=1.23\columnwidth]{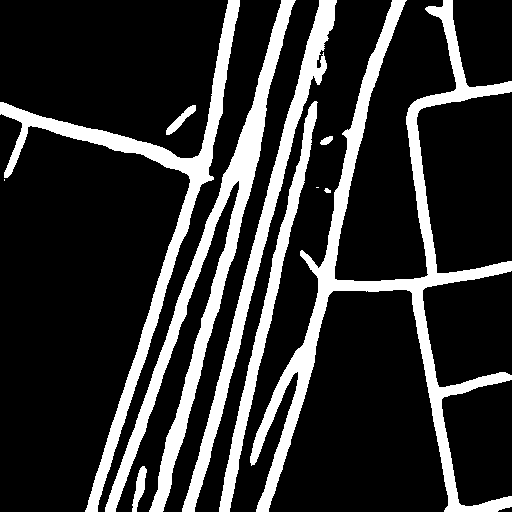} \\
			\vspace{-0.3cm}
			\includegraphics[width=1.23\columnwidth]{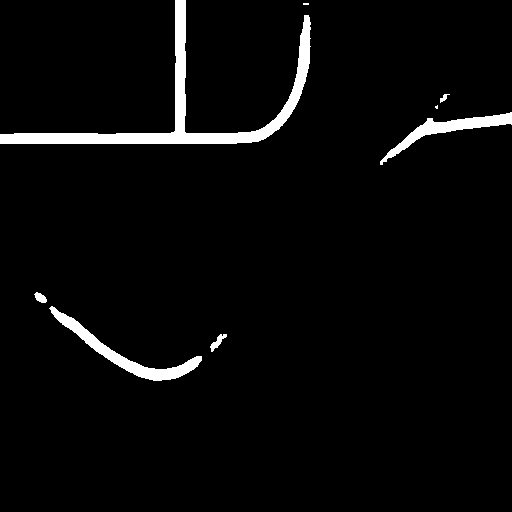} \\
			\vspace{-0.3cm}
			\includegraphics[width=1.23\columnwidth]{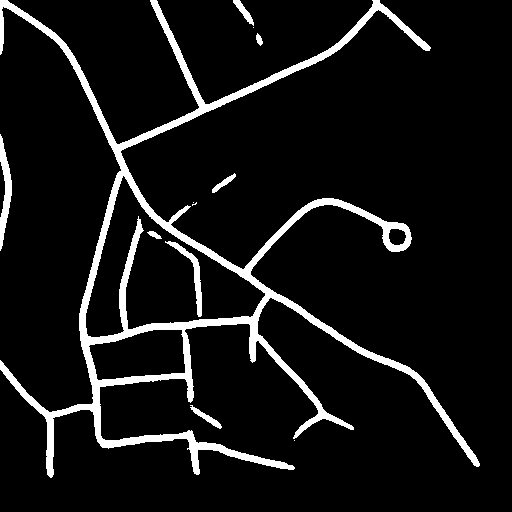} \\
			\vspace{-0.3cm}
			\includegraphics[width=1.23\columnwidth]{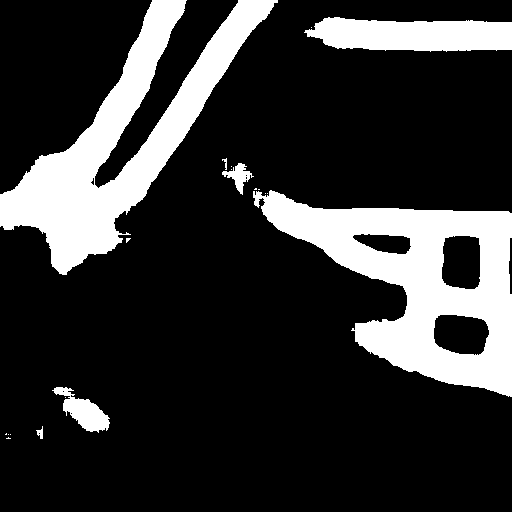} \\
			\vspace{-0.3cm}
			\includegraphics[width=1.23\columnwidth]{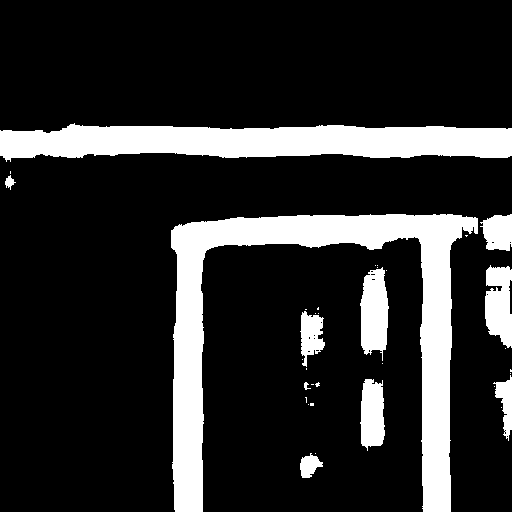} \\
			\vspace{-0.3cm}
			\includegraphics[width=1.23\columnwidth]{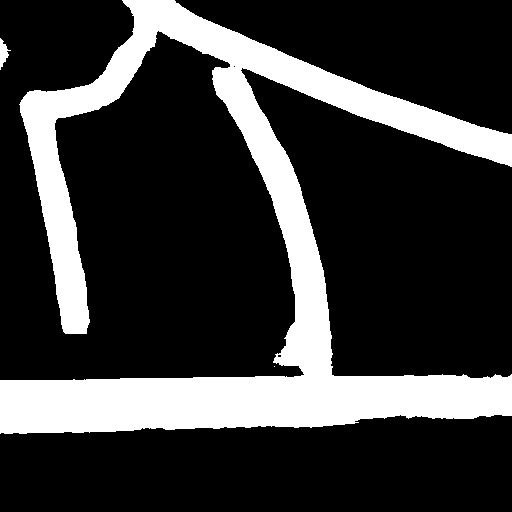}
		\end{minipage}
	}
	\subfigure{
		\begin{minipage}{0.07\textwidth}
			\includegraphics[width=1.23\columnwidth]{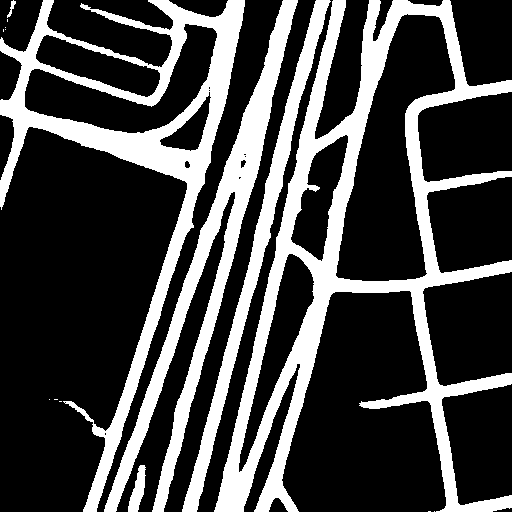} \\
			\vspace{-0.3cm}
			\includegraphics[width=1.23\columnwidth]{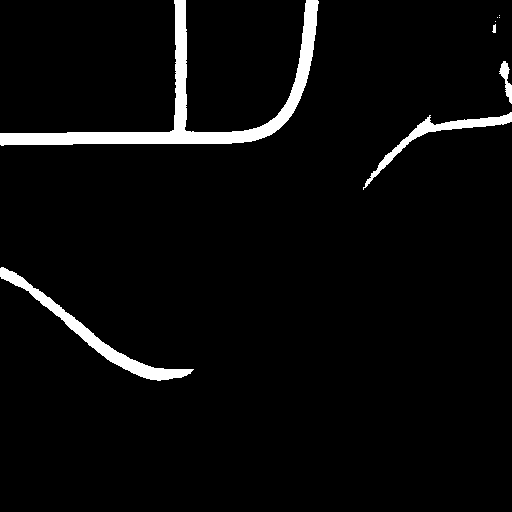} \\
			\vspace{-0.3cm}
			\includegraphics[width=1.23\columnwidth]{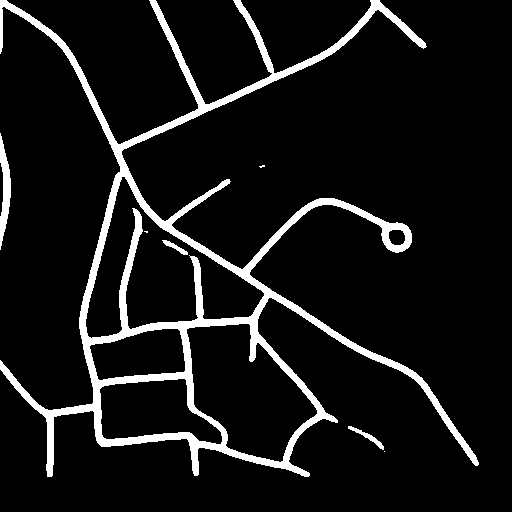} \\
			\vspace{-0.3cm}
			\includegraphics[width=1.23\columnwidth]{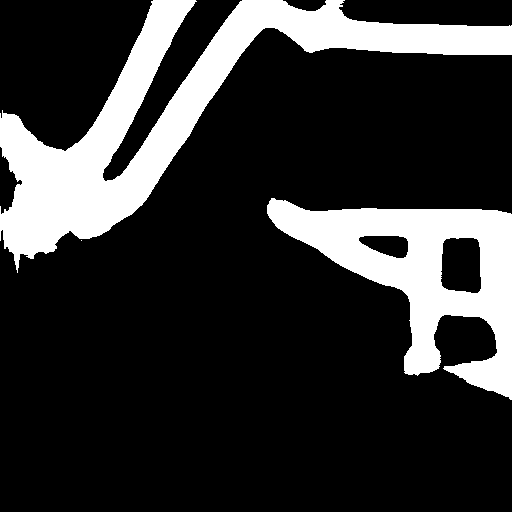} \\
			\vspace{-0.3cm}
			\includegraphics[width=1.23\columnwidth]{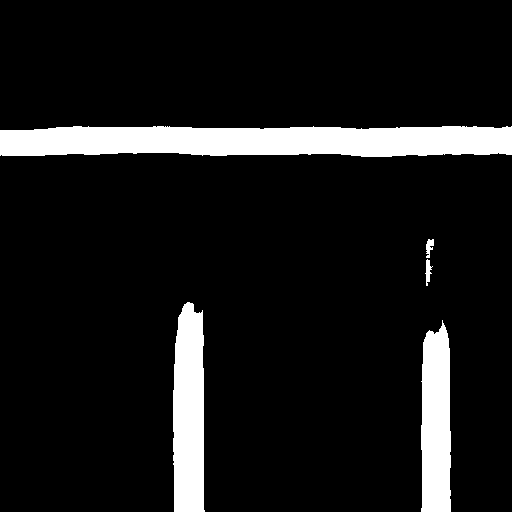} \\
			\vspace{-0.3cm}
			\includegraphics[width=1.23\columnwidth]{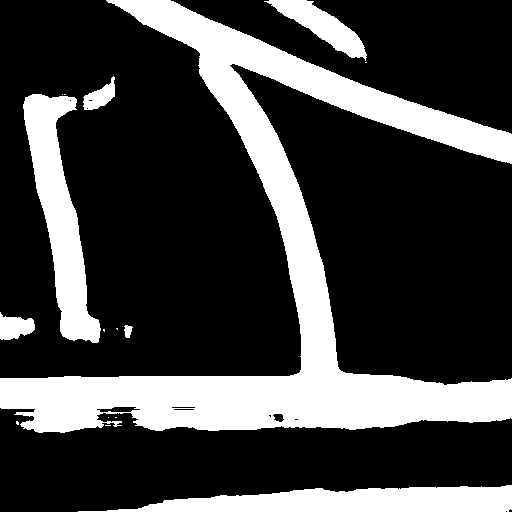}
		\end{minipage}
	}
	\subfigure{
		\begin{minipage}{0.07\textwidth}
			\includegraphics[width=1.23\columnwidth]{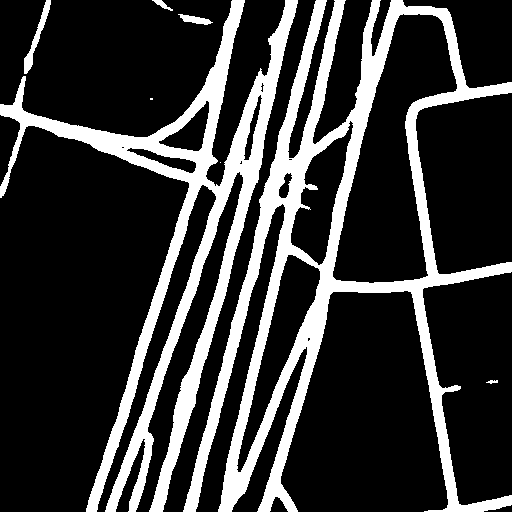} \\
			\vspace{-0.3cm}
			\includegraphics[width=1.23\columnwidth]{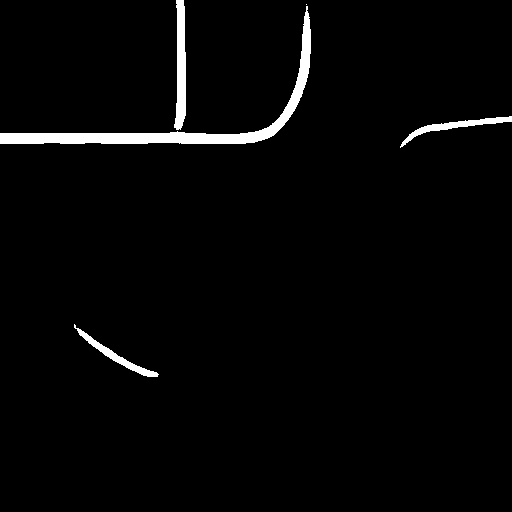} \\
			\vspace{-0.3cm}
			\includegraphics[width=1.23\columnwidth]{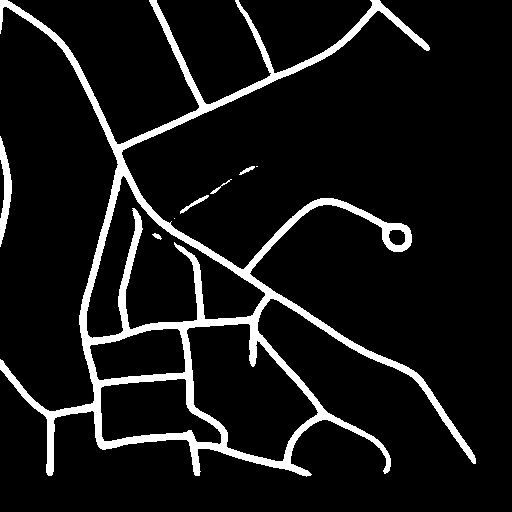} \\
			\vspace{-0.3cm}
			\includegraphics[width=1.23\columnwidth]{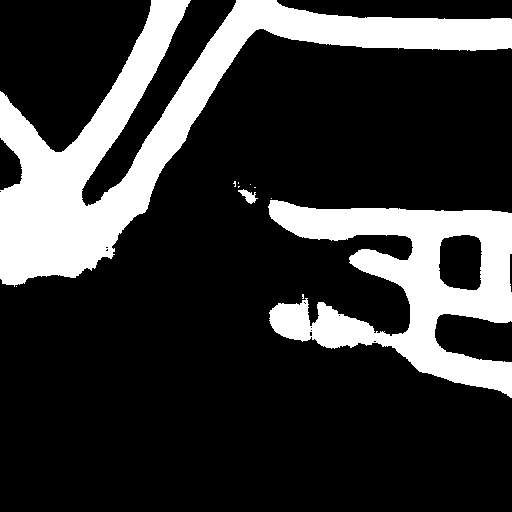} \\
			\vspace{-0.3cm}
			\includegraphics[width=1.23\columnwidth]{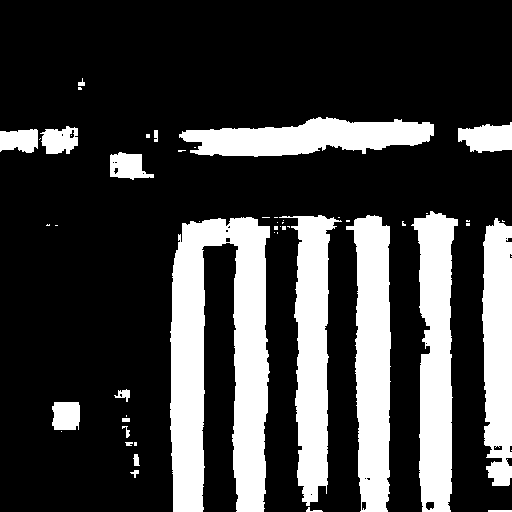} \\
			\vspace{-0.3cm}
			\includegraphics[width=1.23\columnwidth]{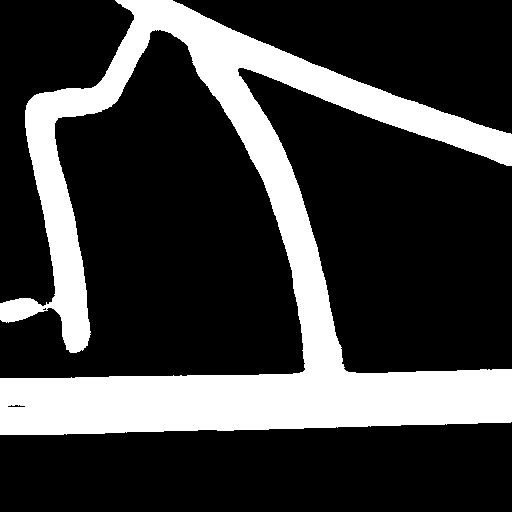}
		\end{minipage}
	}
	\subfigure{
		\begin{minipage}{0.07\textwidth}
			\includegraphics[width=1.23\columnwidth]{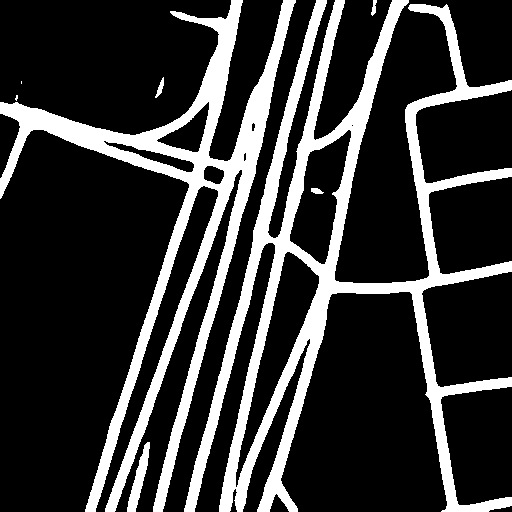} \\
			\vspace{-0.3cm}
			\includegraphics[width=1.23\columnwidth]{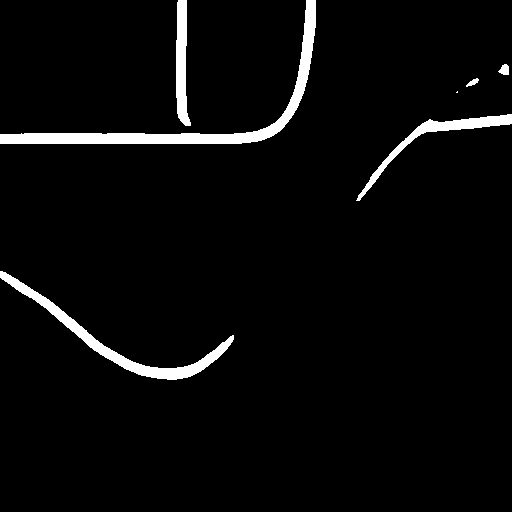} \\
			\vspace{-0.3cm}
			\includegraphics[width=1.23\columnwidth]{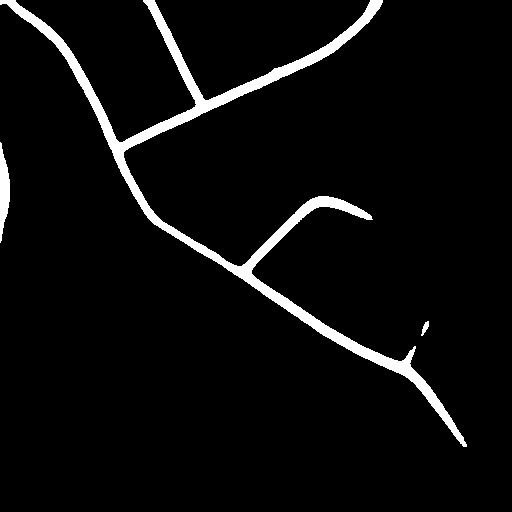} \\
			\vspace{-0.3cm}
			\includegraphics[width=1.23\columnwidth]{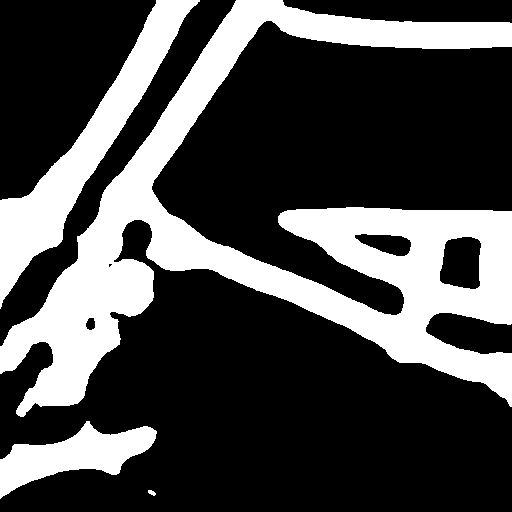} \\
			\vspace{-0.3cm}
			\includegraphics[width=1.23\columnwidth]{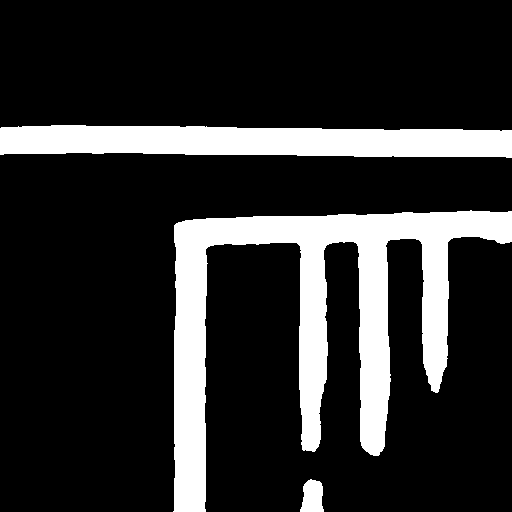} \\
			\vspace{-0.3cm}
			\includegraphics[width=1.23\columnwidth]{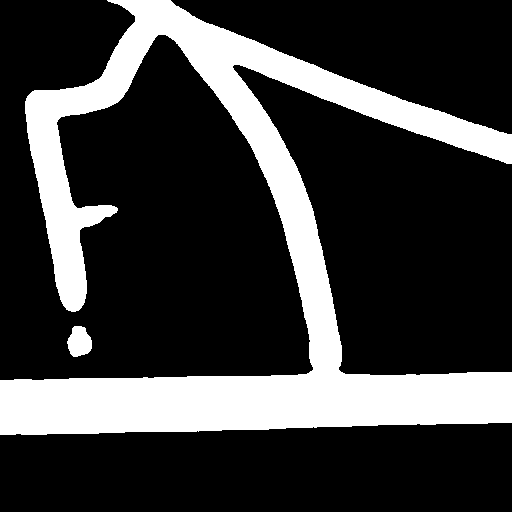}
		\end{minipage}
	}
	\subfigure{
		\begin{minipage}{0.07\textwidth}
			\includegraphics[width=1.23\columnwidth]{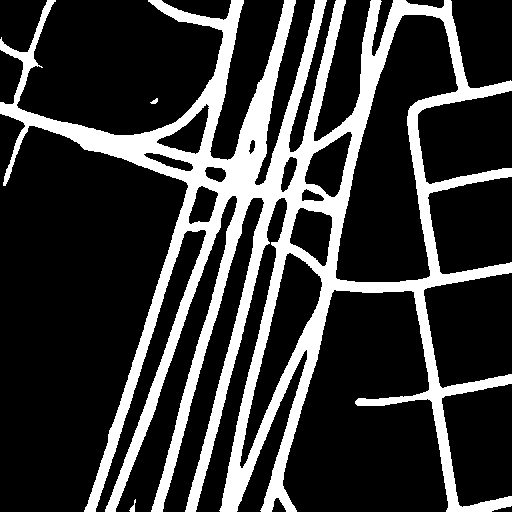} \\
			\vspace{-0.3cm}
			\includegraphics[width=1.23\columnwidth]{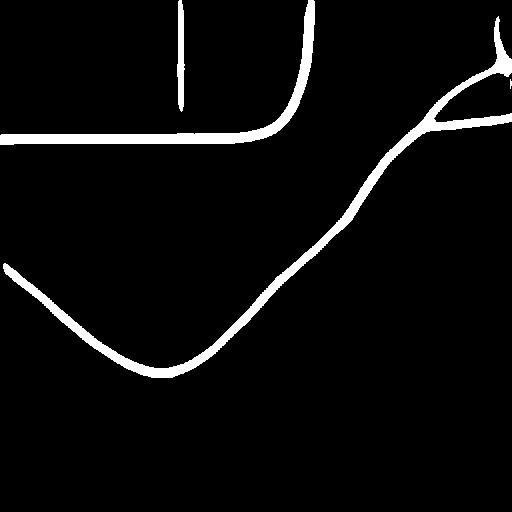} \\
			\vspace{-0.3cm}
			\includegraphics[width=1.23\columnwidth]{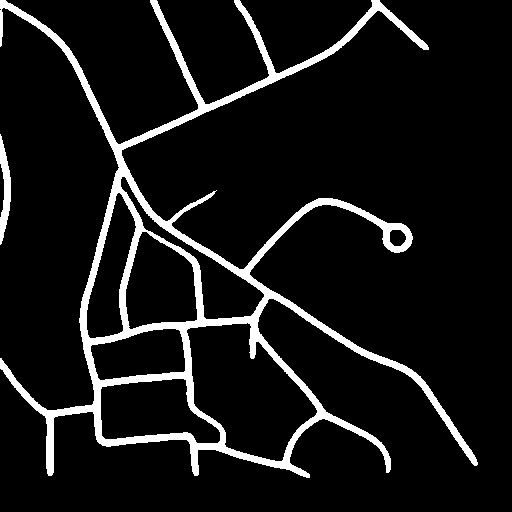} \\
			\vspace{-0.3cm}
			\includegraphics[width=1.23\columnwidth]{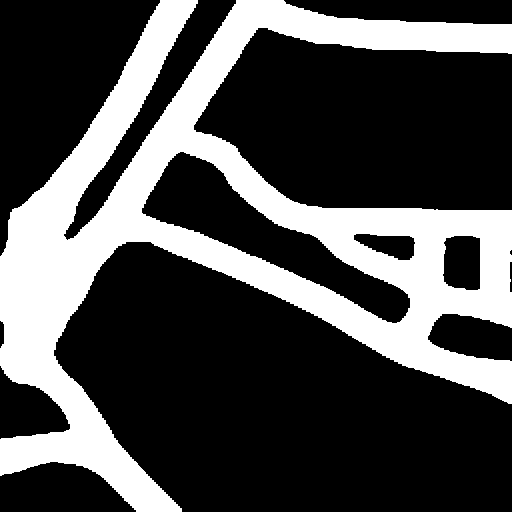} \\
			\vspace{-0.3cm}
			\includegraphics[width=1.23\columnwidth]{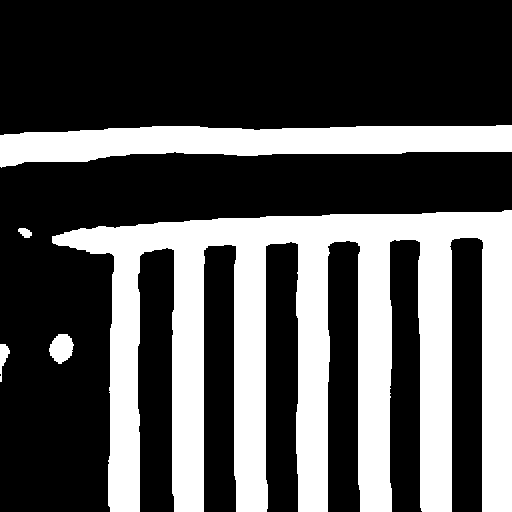} \\
			\vspace{-0.3cm}
			\includegraphics[width=1.23\columnwidth]{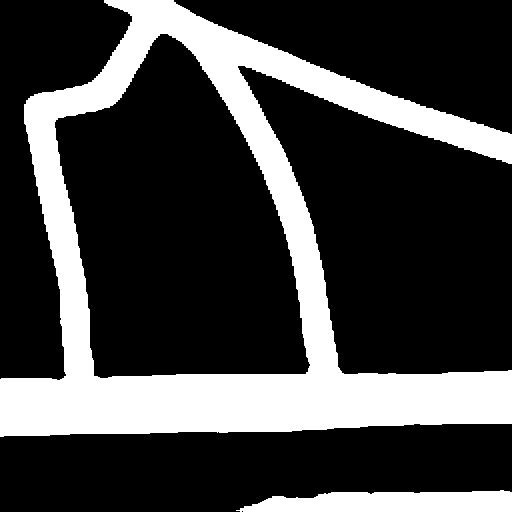}
		\end{minipage}
	}
	\begin{minipage}{1\textwidth}
		\footnotesize
		\hspace{1.5cm}
		Image
		\hspace{0.05\textwidth}
		GT
		\hspace{0.05\textwidth}
		UNet
		\hspace{0.04\textwidth}
		LinkNet
		\hspace{0.04\textwidth}
		DLA
		\hspace{0.04\textwidth}
		DRM
		\hspace{0.04\textwidth}
		Li. \textit{et al.}
		\hspace{0.02\textwidth}
		D-LinkNet
		\hspace{0.02\textwidth}
		B. \textit{et al.}
		\hspace{0.035\textwidth}
		Ours
		\hfil
	\end{minipage}

\caption{Qualitative comparisons to the state-of-the-art methods.}
\label{fig:com}
\end{figure*}

\begin{table*}[!t]
	\renewcommand{\arraystretch}{1.3}
	\caption{Quantitative performance comparisons of our method with the state-of-the-art segmentation methods. Our approach achieves the best results on 3 datasets in terms of IoU and APLS.}
	\label{tb:comparison}
	\centering
	\begin{tabular}{ c | c c c | c c c | c c c }
		\hline
		\multirow{2}*{Methods} & 
		\multicolumn{3}{c|}{\textbf{SpaceNet}} & \multicolumn{3}{c|}{\textbf{RoadTracer}} & \multicolumn{3}{c}{\textbf{Massachusetts}}\\ 
		&IoU  & $\mbox{IoU}^r$ & APLS & IoU & $\mbox{IoU}^r$ & APLS & IoU & $\mbox{IoU}^r$ & APLS   \\ 
		\hline
		UNet & 61.33 & 64.70 & 56.28 & 51.10 & 60.26 & 60.11 & 62.33 & 71.92 & 73.98 \\
		UNet+GA(ours) & 62.67 & 66.14 & 58.26 & 54.03 & 62.35 & 62.16 & 62.75 & 72.43 & 76.84 \\
		LinkNet & 63.61 & 66.44 & 64.84 & 50.73 & 60.22 & 67.00 & 59.30 & 70.20 & 71.38 \\
		LinkNet+connectivity(ours) & 66.14 & 69.79 & 67.82 & 54.92 & 64.60 & 74.49 & 62.57 & 72.64 & 76.60 \\
		DLA & 63.71 & 67.24 & 64.14 & 53.81 & 63.35 & 71.77 & 60.66 & 70.78 & 73.25 \\
		DeepRoadMapper(seg) & 63.49 & 66.98 & 63.52 & 53.73 & 63.44 & 70.44 & 59.91 & 69.66 & 71.13 \\
		Li \textit{et al.} & 63.98 & 66.76 & 66.45 & 53.33 & 65.10 & 72.64 & 61.91 & 73.03 & 76.35 \\
		D-LinkNet & 64.41 & 67.82 & 65.60 & 54.23 & 63.63 & 72.90 & 59.55 & 71.90 & 74.87 \\
		Batra \textit{et al.} & 64.34 & 68.11 & 65.17 & 55.62 & 65.24 & 73.27 & 62.33 & 72.46 & 76.29 \\
		Ours(full) & \textbf{67.51} & \textbf{70.79} & \textbf{68.87} & \textbf{56.63} & \textbf{66.10} & \textbf{76.18} & \textbf{63.62} & \textbf{73.38} & \textbf{77.75}\\
		\hline
	\end{tabular}
\end{table*}

Despite the large difference between the three datasets, especially the exponential increase in ground resolution (0.3 m/pixel, 0.6 m/pixel, and 1.2 m/pixel), our approach significantly improves road segmentation accuracy and road graph connectivity. In Fig. 13, we show comparisons with the above methods in some challenging situations, where our method achieves better results in occlusion and shadow areas, although it also faces challenges in accurately connected roads under these occlusions. Lack of texture and spectral similarity are also the main causes of false alarms. Our method can connect road segments under the overpass in Fig. \ref{fig:com} (a), detect the road segments heavily covered by trees in Fig. \ref{fig:com} (b), identify unsurfaced roads in Fig. \ref{fig:com} (c), and parking lots without cars in (e), which are difficult to distinguish due to lack of texture information. Although our method cannot completely detect the largely shaded roads illustrated in Fig. \ref{fig:com} (d), it does provide good completeness. All the above methods have detected unlabeled roads in Fig. \ref{fig:com}, but our method better preserves the connectivity of the segments.

\section{Conclusion}

In this paper, we argue that geographical scene complexity is the bottleneck of remote sensing imagery interpretation, and road extraction, as a particular interpretation task, usually generates fragmented road segments with poor connectivity. To alleviate these problems, we develop a stacked multihead network to simultaneously perform road segmentation and road connection. In the repeated encoding and decoding process, this network automatically learns and integrates multiscale features to cope with various geographical scenes and road scales, such as urban main roads and outskirt unpaved tracks. Aimed at road elongation and background distraction, we use a global awareness module in residual blocks, which focuses the network on road-related feature expression and reduces the influence of confusing geographic surroundings, thus improving the accuracy of road segmentation. Moreover, we explore the methods to address connectivity issues caused by shadows, occlusion, and homogeneous confusion and develop a novel road-direction-related connectivity task. Joint learning of explicit connectivity labels and road segmentation tasks constrains the model to generate topologically connected road network masks, and therefore, small segmentation cracks can be connected. However, our proposed method faces challenges in predicting roads under large shadow occlusions (in Fig. \ref{fig:com} (d)) and parking many roads that lack texture and contextual features (in Fig. \ref{fig:com} (e)). The comprehensive experimental results show the effectiveness of our network and a significant improvement in road connectivity, even roads under trees and overpasses. It has a stable and outstanding performance in three different datasets, which demonstrates that colearning of the related connectivity task and the road segmentation task can improve the generalization and robustness of the model. Nonetheless, we also observe some poor performances in low-resolution images, where footpaths are occasionally undetected due to road thinness. In the future, we plan to further study road detection using relatively low-resolution images (10 m/pixels or 30 m/pixels), which are more widely used in real-world applications.


%



\section*{Acknowledgment}
	
This work was supported by the National Natural Science Foundation of China under Grant 41925006.
%
%

\bibliography{IEEEabrv, ref/paper}

\begin{thebibliography}{10}
\providecommand{\url}[1]{#1}
\csname url@samestyle\endcsname
\providecommand{\newblock}{\relax}
\providecommand{\bibinfo}[2]{#2}
\providecommand{\BIBentrySTDinterwordspacing}{\spaceskip=0pt\relax}
\providecommand{\BIBentryALTinterwordstretchfactor}{4}
\providecommand{\BIBentryALTinterwordspacing}{\spaceskip=\fontdimen2\font plus
\BIBentryALTinterwordstretchfactor\fontdimen3\font minus \fontdimen4\font\relax}
\providecommand{\BIBforeignlanguage}[2]{{%
\expandafter\ifx\csname l@#1\endcsname\relax
\typeout{** WARNING: IEEEtran.bst: No hyphenation pattern has been}%
\typeout{** loaded for the language `#1'. Using the pattern for}%
\typeout{** the default language instead.}%
\else
\language=\csname l@#1\endcsname
\fi
#2}}
\providecommand{\BIBdecl}{\relax}
\BIBdecl

\bibitem{geraud_fast_2004}
\BIBentryALTinterwordspacing
T.~Géraud and J.-B. Mouret, ``\BIBforeignlanguage{en}{Fast {Road} {Network} {Extraction} in {Satellite} {Images} {Using} {Mathematical} {Morphology} and {Markov} {Random} {Fields}},'' \emph{\BIBforeignlanguage{en}{EURASIP Journal on Advances in Signal Processing}}, vol. 2004, no.~16, p. 473593, Dec. 2004. [Online]. Available: \url{https://doi.org/10.1155/S1110865704409093}
\BIBentrySTDinterwordspacing

\bibitem{cheng_urban_2014}
G.~Cheng, Y.~Wang, Y.~Gong, F.~Zhu, and C.~Pan, ``Urban road extraction via graph cuts based probability propagation,'' in \emph{2014 {IEEE} {International} {Conference} on {Image} {Processing} ({ICIP})}, Oct. 2014, pp. 5072--5076, iSSN: 2381-8549.

\bibitem{doucette_self-organised_2001}
\BIBentryALTinterwordspacing
P.~Doucette, P.~Agouris, A.~Stefanidis, and M.~Musavi, ``\BIBforeignlanguage{en}{Self-organised clustering for road extraction in classified imagery},'' \emph{\BIBforeignlanguage{en}{ISPRS Journal of Photogrammetry and Remote Sensing}}, vol.~55, no.~5, pp. 347--358, Mar. 2001. [Online]. Available: \url{http://www.sciencedirect.com/science/article/pii/S0924271601000272}
\BIBentrySTDinterwordspacing

\bibitem{cohen_global_1996}
\BIBentryALTinterwordspacing
L.~Cohen and R.~Kimmel, ``\BIBforeignlanguage{en}{Global minimum for active contour models: a minimal path approach},'' in \emph{\BIBforeignlanguage{en}{Proceedings {CVPR} {IEEE} {Computer} {Society} {Conference} on {Computer} {Vision} and {Pattern} {Recognition}}}.\hskip 1em plus 0.5em minus 0.4em\relax San Francisco, CA, USA: IEEE, 1996, pp. 666--673. [Online]. Available: \url{http://ieeexplore.ieee.org/document/517144/}
\BIBentrySTDinterwordspacing

\bibitem{kass_snakes_1988}
\BIBentryALTinterwordspacing
M.~Kass, A.~Witkin, and D.~Terzopoulos, ``\BIBforeignlanguage{en}{Snakes: {Active} contour models},'' \emph{\BIBforeignlanguage{en}{International Journal of Computer Vision}}, vol.~1, no.~4, pp. 321--331, Jan. 1988. [Online]. Available: \url{https://doi.org/10.1007/BF00133570}
\BIBentrySTDinterwordspacing

\bibitem{mattyus_deeproadmapper_2017}
G.~Máttyus, W.~Luo, and R.~Urtasun, ``{DeepRoadMapper}: {Extracting} {Road} {Topology} from {Aerial} {Images},'' in \emph{2017 {IEEE} {International} {Conference} on {Computer} {Vision} ({ICCV})}, 2017, pp. 3458--3466.

\bibitem{bastani_roadtracer:_2018}
\BIBentryALTinterwordspacing
F.~Bastani, S.~He, S.~Abbar, M.~Alizadeh, H.~Balakrishnan, S.~Chawla, S.~Madden, and D.~DeWitt, ``\BIBforeignlanguage{en}{{RoadTracer}: {Automatic} {Extraction} of {Road} {Networks} from {Aerial} {Images}},'' in \emph{\BIBforeignlanguage{en}{2018 {IEEE}/{CVF} {Conference} on {Computer} {Vision} and {Pattern} {Recognition}}}.\hskip 1em plus 0.5em minus 0.4em\relax Salt Lake City, UT: IEEE, Jun. 2018, pp. 4720--4728, cVPR. [Online]. Available: \url{https://ieeexplore.ieee.org/document/8578594/}
\BIBentrySTDinterwordspacing

\bibitem{ventura_iterative_2018}
\BIBentryALTinterwordspacing
C.~Ventura, J.~Pont-Tuset, S.~Caelles, K.-K. Maninis, and L.~Van~Gool, ``\BIBforeignlanguage{en}{Iterative {Deep} {Learning} for {Road} {Topology} {Extraction}},'' \emph{\BIBforeignlanguage{en}{arXiv:1808.09814 [cs]}}, Aug. 2018. [Online]. Available: \url{http://arxiv.org/abs/1808.09814}
\BIBentrySTDinterwordspacing

\bibitem{li_topological_2019}
\BIBentryALTinterwordspacing
Z.~Li, J.~D. Wegner, and A.~Lucchi, ``\BIBforeignlanguage{en}{Topological {Map} {Extraction} from {Overhead} {Images}},'' \emph{\BIBforeignlanguage{en}{arXiv:1812.01497 [cs]}}, Apr. 2019, arXiv: 1812.01497. [Online]. Available: \url{http://arxiv.org/abs/1812.01497}
\BIBentrySTDinterwordspacing

\bibitem{tan_vecroad_2020}
\BIBentryALTinterwordspacing
Y.-Q. Tan, S.-H. Gao, X.-Y. Li, M.-M. Cheng, and B.~Ren, ``\BIBforeignlanguage{en}{{VecRoad}: {Point}-{Based} {Iterative} {Graph} {Exploration} for {Road} {Graphs} {Extraction}},'' in \emph{\BIBforeignlanguage{en}{2020 {IEEE}/{CVF} {Conference} on {Computer} {Vision} and {Pattern} {Recognition} ({CVPR})}}.\hskip 1em plus 0.5em minus 0.4em\relax Seattle, WA, USA: IEEE, Jun. 2020, pp. 8907--8915, cVPR. [Online]. Available: \url{https://ieeexplore.ieee.org/document/9157398/}
\BIBentrySTDinterwordspacing

\bibitem{li_topology-enhanced_2020}
X.~Li, Y.~Wang, L.~Zhang, S.~Liu, J.~Mei, and Y.~Li, ``Topology-{Enhanced} {Urban} {Road} {Extraction} via a {Geographic} {Feature}-{Enhanced} {Network},'' \emph{IEEE Transactions on Geoscience and Remote Sensing}, pp. 1--12, 2020, liXG.

\bibitem{newell_stacked_2016}
A.~Newell, K.~Yang, and J.~Deng, ``Stacked hourglass networks for human pose estimation,'' in \emph{European conference on computer vision}.\hskip 1em plus 0.5em minus 0.4em\relax Springer, 2016, pp. 483--499.

\bibitem{steger_model-based_1995}
C.~Steger, C.~Glock, W.~Eckstein, H.~Mayer, and B.~Radig, ``Model-{Based} {Road} {Extraction} from {Images},'' in \emph{In: {Automatic} {Extraction} of {Man}-{Made} {Objects} from {Aerial} and {Space} {Images}, {Birkh} auser {Verlag} {Basel}}.\hskip 1em plus 0.5em minus 0.4em\relax Birkhauser Verlag, 1995, pp. 275--284.

\bibitem{treash_automatic_2001}
K.~Treash, ``Automatic {Road} {Detection} {In} {Grayscale} {Aerial} {Images},'' \emph{Journal of Computing in Civil Engineering}, vol.~14, Jul. 2001.

\bibitem{ma_road_2007}
H.~Ma, Q.~Qin, S.~Du, L.~Wang, and C.~Jin, ``Road extraction from {ETM} panchromatic image based on {Dual}-{Edge} {Following},'' Aug. 2007, pp. 460--463.

\bibitem{sengupta_phase-based_nodate}
\BIBentryALTinterwordspacing
S.~K. Sengupta, A.~S. Lopez, J.~M. Brase, and D.~W. Paglieroni, \emph{\BIBforeignlanguage{English}{Phase-based road detection in multi-source images}}, publication Title: IGARSS 2004. 2004 IEEE International Geoscience and Remote Sensing Symposium. [Online]. Available: \url{https://www.infona.pl//resource/bwmeta1.element.ieee-art-000001369959}
\BIBentrySTDinterwordspacing

\bibitem{yager_support_2003}
N.~Yager and A.~Sowmya, ``Support {Vector} {Machines} for {Road} {Extraction} from {Remotely} {Sensed} {Images},'' Aug. 2003, pp. 285--292.

\bibitem{doucette_self-organised_2001-1}
\BIBentryALTinterwordspacing
P.~Doucette, P.~Agouris, A.~Stefanidis, and M.~Musavi, ``\BIBforeignlanguage{en}{Self-organised clustering for road extraction in classified imagery},'' \emph{\BIBforeignlanguage{en}{ISPRS Journal of Photogrammetry and Remote Sensing}}, vol.~55, no.~5, pp. 347--358, Mar. 2001. [Online]. Available: \url{http://www.sciencedirect.com/science/article/pii/S0924271601000272}
\BIBentrySTDinterwordspacing

\bibitem{park_semi-automatic_2001}
S.-R. Park and T.~Kim, ``Semi-automatic road extraction algorithm from {IKONOS} images using template matching,'' in \emph{Proc. 22nd {Asian} {Conference} on {Remote} {Sensing}}, 2001, pp. 1209--1213.

\bibitem{chaudhuri_semi-automated_2012-1}
D.~Chaudhuri, N.~K. Kushwaha, and A.~Samal, ``Semi-{Automated} {Road} {Detection} {From} {High} {Resolution} {Satellite} {Images} by {Directional} {Morphological} {Enhancement} and {Segmentation} {Techniques},'' \emph{IEEE Journal of Selected Topics in Applied Earth Observations and Remote Sensing}, vol.~5, no.~5, pp. 1538--1544, Oct. 2012, conference Name: IEEE Journal of Selected Topics in Applied Earth Observations and Remote Sensing.

\bibitem{shi_integrated_2014}
W.~Shi, Z.~Miao, and J.~Debayle, ``An {Integrated} {Method} for {Urban} {Main}-{Road} {Centerline} {Extraction} {From} {Optical} {Remotely} {Sensed} {Imagery},'' \emph{IEEE Transactions on Geoscience and Remote Sensing}, vol.~52, no.~6, pp. 3359--3372, Jun. 2014, conference Name: IEEE Transactions on Geoscience and Remote Sensing.

\bibitem{zhou_d-linknet:_2018}
\BIBentryALTinterwordspacing
L.~Zhou, C.~Zhang, and M.~Wu, ``\BIBforeignlanguage{en}{D-{LinkNet}: {LinkNet} with {Pretrained} {Encoder} and {Dilated} {Convolution} for {High} {Resolution} {Satellite} {Imagery} {Road} {Extraction}},'' in \emph{\BIBforeignlanguage{en}{2018 {IEEE}/{CVF} {Conference} on {Computer} {Vision} and {Pattern} {Recognition} {Workshops} ({CVPRW})}}.\hskip 1em plus 0.5em minus 0.4em\relax Salt Lake City, UT, USA: IEEE, Jun. 2018, pp. 192--1924. [Online]. Available: \url{https://ieeexplore.ieee.org/document/8575492/}
\BIBentrySTDinterwordspacing

\bibitem{tao_spatial_2019}
\BIBentryALTinterwordspacing
C.~Tao, J.~Qi, Y.~Li, H.~Wang, and H.~Li, ``\BIBforeignlanguage{en}{Spatial information inference net: {Road} extraction using road-specific contextual information},'' \emph{\BIBforeignlanguage{en}{ISPRS Journal of Photogrammetry and Remote Sensing}}, vol. 158, pp. 155--166, Dec. 2019. [Online]. Available: \url{http://www.sciencedirect.com/science/article/pii/S0924271619302382}
\BIBentrySTDinterwordspacing

\bibitem{liu_d-resunet_2019}
Z.~Liu, R.~Feng, L.~Wang, Y.~Zhong, and L.~Cao, ``D-{Resunet}: {Resunet} and {Dilated} {Convolution} for {High} {Resolution} {Satellite} {Imagery} {Road} {Extraction},'' in \emph{{IGARSS} 2019 - 2019 {IEEE} {International} {Geoscience} and {Remote} {Sensing} {Symposium}}, Jul. 2019, pp. 3927--3930, iSSN: 2153-7003.

\bibitem{sun_stacked_2018}
T.~Sun, Z.~Chen, W.~Yang, and Y.~Wang, ``Stacked {U}-{Nets} {With} {Multi}-{Output} for {Road} {Extraction},'' in \emph{The {IEEE} {Conference} on {Computer} {Vision} and {Pattern} {Recognition} ({CVPR}) {Workshops}}, Jun. 2018, cVPR.

\bibitem{mnih_learning_2010}
V.~Mnih and G.~E. Hinton, ``\BIBforeignlanguage{en}{Learning to {Detect} {Roads} in {High}-{Resolution} {Aerial} {Images}},'' in \emph{\BIBforeignlanguage{en}{Computer {Vision} – {ECCV} 2010}}, ser. Lecture {Notes} in {Computer} {Science}, K.~Daniilidis, P.~Maragos, and N.~Paragios, Eds.\hskip 1em plus 0.5em minus 0.4em\relax Berlin, Heidelberg: Springer, 2010, pp. 210--223.

\bibitem{li_road_2016}
P.~Li, Y.~Zang, C.~Wang, J.~Li, M.~Cheng, L.~Luo, and Y.~Yu, ``Road network extraction via deep learning and line integral convolution,'' in \emph{2016 {IEEE} {International} {Geoscience} and {Remote} {Sensing} {Symposium} ({IGARSS})}, Jul. 2016, pp. 1599--1602, iSSN: 2153-7003.

\bibitem{mosinska_beyond_2018}
A.~Mosinska, P.~Marquez-Neila, M.~Kozi{\'n}ski, and P.~Fua, ``Beyond the pixel-wise loss for topology-aware delineation,'' in \emph{Proceedings of the IEEE conference on computer vision and pattern recognition}, 2018, pp. 3136--3145.

\bibitem{batra_improved_2019}
\BIBentryALTinterwordspacing
A.~Batra, S.~Singh, G.~Pang, S.~Basu, C.~Jawahar, and M.~Paluri, ``\BIBforeignlanguage{en}{Improved {Road} {Connectivity} by {Joint} {Learning} of {Orientation} and {Segmentation}},'' in \emph{\BIBforeignlanguage{en}{2019 {IEEE}/{CVF} {Conference} on {Computer} {Vision} and {Pattern} {Recognition} ({CVPR})}}.\hskip 1em plus 0.5em minus 0.4em\relax Long Beach, CA, USA: IEEE, Jun. 2019, pp. 10\,377--10\,385, orientation and segmentation, spacent. [Online]. Available: \url{https://ieeexplore.ieee.org/document/8953380/}
\BIBentrySTDinterwordspacing

\bibitem{yu_casenet:_2017}
Z.~Yu, C.~Feng, M.-Y. Liu, and S.~Ramalingam, ``Casenet: {Deep} category-aware semantic edge detection,'' in \emph{Proceedings of the {IEEE} {Conference} on {Computer} {Vision} and {Pattern} {Recognition}}, 2017, pp. 5964--5973, caseNet(CVPR).

\bibitem{liu_roadnet:_2019}
Y.~Liu, J.~Yao, X.~Lu, M.~Xia, X.~Wang, and Y.~Liu, ``\BIBforeignlanguage{en}{{RoadNet}: {Learning} to {Comprehensively} {Analyze} {Road} {Networks} in {Complex} {Urban} {Scenes} {From} {High}-{Resolution} {Remotely} {Sensed} {Images}},'' \emph{\BIBforeignlanguage{en}{IEEE TRANSACTIONS ON GEOSCIENCE AND REMOTE SENSING}}, vol.~57, no.~4, p.~14, 2019.

\bibitem{ronneberger_u-net:_2015}
O.~Ronneberger, P.~Fischer, and T.~Brox, ``U-net: Convolutional networks for biomedical image segmentation,'' in \emph{International Conference on Medical image computing and computer-assisted intervention}.\hskip 1em plus 0.5em minus 0.4em\relax Springer, 2015, pp. 234--241.

\bibitem{li_feature_2019}
T.~Li, M.~Comer, and J.~Zerubia, ``Feature {Extraction} and {Tracking} of {CNN} {Segmentations} for {Improved} {Road} {Detection} from {Satellite} {Imagery},'' in \emph{2019 {IEEE} {International} {Conference} on {Image} {Processing} ({ICIP})}, Sep. 2019, pp. 2641--2645, iSSN: 2381-8549.

\bibitem{buslaev_fully_2018}
A.~Buslaev, S.~S. Seferbekov, V.~Iglovikov, and A.~Shvets, ``Fully {Convolutional} {Network} for {Automatic} {Road} {Extraction} {From} {Satellite} {Imagery}.'' in \emph{{CVPR} {Workshops}}, 2018, pp. 207--210.

\bibitem{filin_road_2018}
O.~Filin, A.~Zapara, and S.~Panchenko, ``Road {Detection} {With} {EOSResUNet} and {Post} {Vectorizing} {Algorithm},'' in \emph{The {IEEE} {Conference} on {Computer} {Vision} and {Pattern} {Recognition} ({CVPR}) {Workshops}}, Jun. 2018, cVPR.

\bibitem{he_deep_2016}
K.~He, X.~Zhang, S.~Ren, and J.~Sun, ``Deep residual learning for image recognition,'' in \emph{Proceedings of the IEEE conference on computer vision and pattern recognition}, 2016, pp. 770--778.

\bibitem{costea_roadmap_2018}
D.~Costea, A.~Marcu, E.~Slusanschi, and M.~Leordeanu, ``Roadmap {Generation} {Using} a {Multi}-{Stage} {Ensemble} of {Deep} {Neural} {Networks} {With} {Smoothing}-{Based} {Optimization},'' in \emph{The {IEEE} {Conference} on {Computer} {Vision} and {Pattern} {Recognition} ({CVPR}) {Workshops}}, Jun. 2018, cVPR.

\bibitem{chaurasia--linknet_2017}
A.~Chaurasia and E.~Culurciello, ``Linknet: Exploiting encoder representations for efficient semantic segmentation,'' in \emph{2017 IEEE Visual Communications and Image Processing (VCIP)}.\hskip 1em plus 0.5em minus 0.4em\relax IEEE, 2017, pp. 1--4.

\bibitem{sun_road_2019}
S.~Sun, W.~Xia, B.~Zhang, and Y.~Zhang, ``Road {Centerlines} {Extraction} from {High} {Resolution} {Remote} {Sensing} {Image},'' in \emph{{IGARSS} 2019 - 2019 {IEEE} {International} {Geoscience} and {Remote} {Sensing} {Symposium}}, Jul. 2019, pp. 3931--3934, iSSN: 2153-7003.

\bibitem{yujun_end--end_2018}
W.~Yujun, H.~Xiangyun, and G.~Jinqi, ``End-to-{End} {Road} {Centerline} {Extraction} via {Learning} a {Confidence} {Map},'' in \emph{2018 10th {IAPR} {Workshop} on {Pattern} {Recognition} in {Remote} {Sensing} ({PRRS})}, Aug. 2018, pp. 1--5.

\bibitem{hu_squeeze-and-excitation_2019}
J.~Hu, L.~Shen, and G.~Sun, ``Squeeze-and-excitation networks,'' in \emph{Proceedings of the IEEE conference on computer vision and pattern recognition}, 2018, pp. 7132--7141.

\bibitem{mnih_machine_2013}
V.~Mnih, ``Machine learning for aerial image labeling,'' {PhD} {Thesis}, University of Toronto, 2013, university of Toronto.

\bibitem{kampffmeyer_connnet_2019}
M.~Kampffmeyer, N.~Dong, X.~Liang, Y.~Zhang, and E.~P. Xing, ``Connnet: A long-range relation-aware pixel-connectivity network for salient segmentation,'' \emph{IEEE Transactions on Image Processing}, vol.~28, no.~5, pp. 2518--2529, 2018.

\bibitem{lin_feature_2017}
T.-Y. Lin, P.~Doll{\'a}r, R.~Girshick, K.~He, B.~Hariharan, and S.~Belongie, ``Feature pyramid networks for object detection,'' in \emph{Proceedings of the IEEE conference on computer vision and pattern recognition}, 2017, pp. 2117--2125.

\bibitem{paszke_enet_2016}
A.~Paszke, A.~Chaurasia, S.~Kim, and E.~Culurciello, ``Enet: A deep neural network architecture for real-time semantic segmentation,'' \emph{arXiv preprint arXiv:1606.02147}, 2016.

\bibitem{van_etten_spacenet:_2019}
\BIBentryALTinterwordspacing
A.~Van~Etten, D.~Lindenbaum, and T.~M. Bacastow, ``\BIBforeignlanguage{en}{{SpaceNet}: {A} {Remote} {Sensing} {Dataset} and {Challenge} {Series}},'' \emph{\BIBforeignlanguage{en}{arXiv:1807.01232 [cs]}}, Jul. 2019, arXiv: 1807.01232. [Online]. Available: \url{http://arxiv.org/abs/1807.01232}
\BIBentrySTDinterwordspacing

\bibitem{noauthor_openstreetmap_nodate}
\BIBentryALTinterwordspacing
``{OpenStreetMap}.'' [Online]. Available: \url{https://www.openstreetmap.org/}
\BIBentrySTDinterwordspacing

\bibitem{yu_deep_2019}
F.~Yu, D.~Wang, E.~Shelhamer, and T.~Darrell, ``Deep layer aggregation,'' in \emph{Proceedings of the IEEE conference on computer vision and pattern recognition}, 2018, pp. 2403--2412.

\end{thebibliography}

\end{document}